%% file: acl_latex.tex
\pdfoutput=1

\documentclass[11pt]{article}
\usepackage[]{acl}
\usepackage{times}
\usepackage{latexsym}
\usepackage[T1]{fontenc}
\usepackage[utf8]{inputenc}
\usepackage{microtype}
\usepackage{inconsolata}
\usepackage{ragged2e}
\usepackage{graphicx}
\usepackage{booktabs}
\usepackage{subcaption}
\usepackage[frozencache,cachedir=.]{minted}
\usepackage{tabularx}
\usepackage{multirow}
\usepackage{tikz}
\usepackage{enumitem}
\usepackage[framemethod=TikZ]{mdframed}

\newmdenv[%
    backgroundcolor=gray!10,
    linecolor=black,
    outerlinewidth=0.5pt,
    roundcorner=1mm,
    skipabove=\topsep,
    skipbelow=\topsep,
    font=\ttfamily\tiny,
]{promptbox}

\newcommand{\systemname}{\textsc{Metal}}

\title{\systemname: Towards Multilingual Meta-Evaluation}
\input{content/authors}
\begin{document}
\maketitle

\input{content/abstract}
\input{content/introduction}
\input{content/related_works}
\input{content/data}
\input{content/experiments}
\input{content/results_discussions}
\input{content/qualitative_analysis}
\input{content/conclusion}
\input{content/limitations}
\input{content/ethical_considerations}
\bibliography{anthology,custom}

\newpage
\appendix
\section{Appendix}
\label{sec:appendix}
\input{appendix/main_prompt}
\input{appendix/task_instructions}
\input{appendix/annotator_agreement}
\input{appendix/simple_instructions}
\input{appendix/F1_scores}
\input{appendix/appendix_qt}
\end{document}

%% file: content/authors.tex
\author{\parbox{0.9\linewidth}{\centering{Rishav Hada\thanks{\hspace{0.1cm} Equal contribution} \quad Varun Gumma\footnotemark[1] \\\quad Mohamed Ahmed  \quad Kalika Bali \quad Sunayana Sitaram \\
Microsoft Corporation \\
\small{{\tt \{rishavhada,varun230999\}@gmail.com, \{maxamed.axmed,kalikab,sunayana.sitaram\}@microsoft.com}} }}}

%% file: content/abstract.tex
\begin{abstract}
With the rising human-like precision of Large Language Models (LLMs) in numerous tasks, their utilization in a variety of real-world applications is becoming more prevalent. Several studies have shown that LLMs excel on many standard NLP benchmarks. However, it is challenging to evaluate LLMs due to test dataset  contamination and the limitations of traditional metrics. Since human evaluations are difficult to collect, there is a growing interest in the community to use LLMs themselves as reference-free evaluators for subjective metrics. However, past work has shown that LLM-based evaluators can exhibit bias and have poor alignment with human judgments. In this study, we propose a framework for an end-to-end assessment of LLMs as evaluators in multilingual scenarios. We create a carefully curated dataset, covering 10 languages containing native speaker judgments for the task of summarization. This dataset is created specifically to evaluate LLM-based evaluators, which we refer to as meta-evaluation (\systemname). We compare the performance of LLM-based evaluators created using GPT-3.5-Turbo, GPT-4, and PaLM2. Our results indicate that LLM-based evaluators based on GPT-4 perform the best across languages, while GPT-3.5-Turbo performs poorly. Additionally, we perform an analysis of the reasoning provided by LLM-based evaluators and find that it often does not match the reasoning provided by human judges.
\end{abstract}

%% file: content/introduction.tex
\section{Introduction}

Recent Large Language Models (LLMs) like GPT-4 \cite{openai2023gpt4}, GPT-3.5-Turbo \cite{ouyang2022training}, PaLM2 \cite{anil2023palm}, Gemini-1.5 \cite{reid2024gemini}, Mistral \cite{jiang2023mistral,jiang2024mixtral} \textit{etc.} have shown impressive performance on a variety of standard NLP tasks across languages \cite{ahuja-etal-2023-mega, ahuja2023megaverse,arora-etal-2023-llms,laskar-etal-2023-systematic,tam-etal-2023-evaluating,zhang2023benchmarking}. However, there are several challenges in fair and accurate assessment of these models, such as the contamination of existing datasets in LLM pre-training data, lack of multilingual datasets \cite{ahuja-etal-2022-beyond}, lack of benchmarks that represent real-world usage of these models, lack of frameworks for consistent subjective evaluations, and budget and access issues for native speaker evaluation. Therefore, there is a growing need for frameworks and resources that address the above challenges and allow us to systematically evaluate LLMs across several dimensions and languages. 

\begin{figure}
    \centering
    \includegraphics[width=0.5\textwidth]{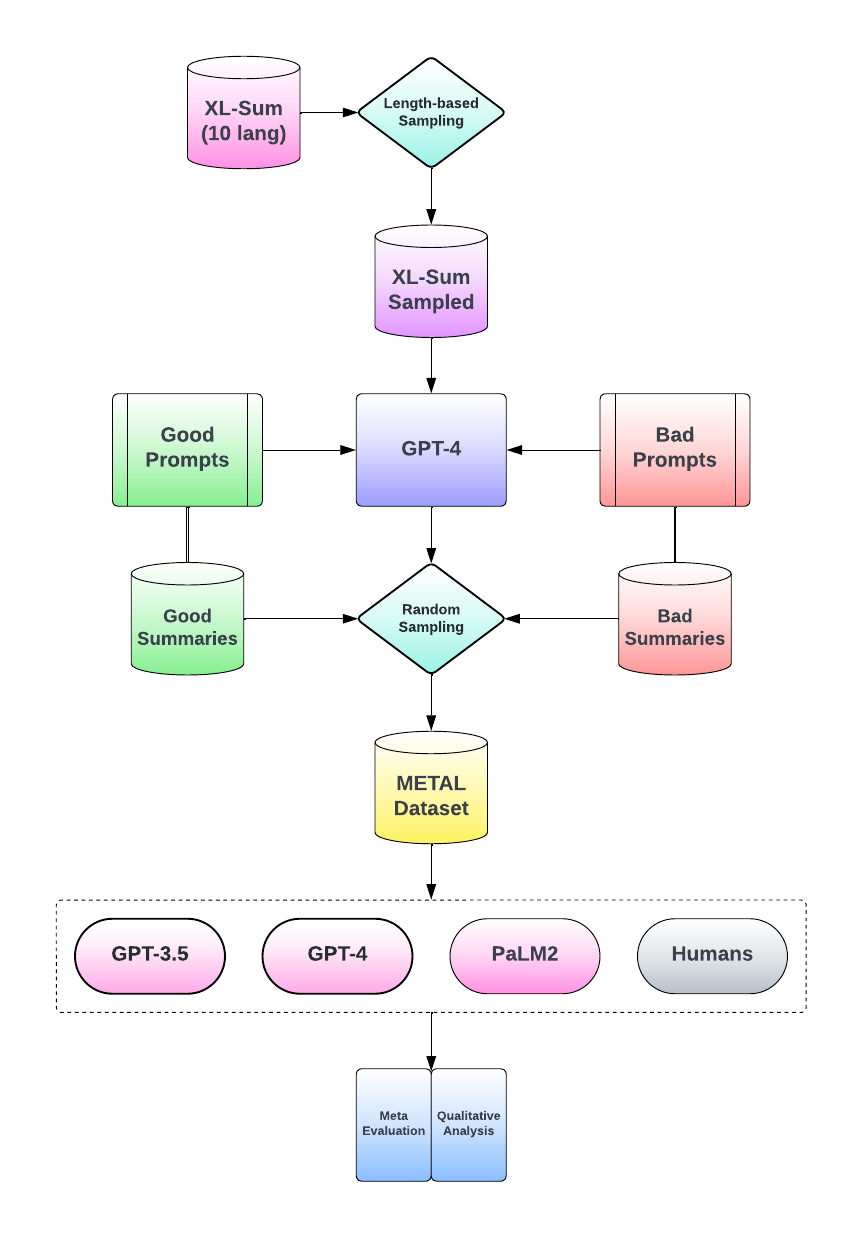}
    \caption{Pipeline of \systemname\ framework.}
    \label{fig:title_figure}
\end{figure}

Further, evaluating the text generation capabilities of these models is even more challenging \cite{chiang-lee-2023-large,zhou-etal-2022-deconstructing,wang-etal-2023-chatgpt}. Natural Language Generation (NLG) capabilities are traditionally evaluated using automated metrics such as ROUGE \cite{lin-2004-rouge} or BLEU \cite{papineni-etal-2002-bleu} scores. These metrics have several known drawbacks, such as reliance on exact matches and over-emphasis on length. Secondly, these metrics do not account for subjective evaluations such as quality, coverage, and coherence \cite{schluter-2017-limits,grusky-2023-rogue,reiter-2018-structured}. Thirdly, these metrics are reference-based i.e. they need a comparison baseline, which can be expensive to collect and can sometimes have a low correlation with human judgments. This has led to work on reference-free and subjective evaluation \cite{chen2023exploring,stammbach-etal-2023-revisiting,xu-etal-2023-instructscore, hasan-etal-2021-xl}. 

Using LLMs as evaluators presents several challenges. Recent works \cite{clark-etal-2021-thats,wang2023large,chiang-lee-2023-large} have shown that while LLMs can produce evaluations with human-like accuracy, these evaluations are often inconsistent and can easily be influenced. LLMs also show position bias or scale region bias and are unable to distinguish between candidates that are close to each other \cite{golchin2023data}. LLMs are sensitive to instructions and their capabilities vary for different metrics \cite{skopek-etal-2023-towards, wang2023large, shen2023large}. Another significant challenge when using LLMs as evaluators is a limited assessment of their abilities in multilingual settings. Studies have shown that LLMs have inferior performance even on some high-resource languages and cannot be assessed extensively on low-resource languages due to a lack of benchmarks \cite{ahuja-etal-2023-mega}. Therefore, it is still unclear if LLMs can replace human evaluations in multilingual settings.

In this paper, we introduce the \systemname\ framework for a robust assessment of LLMs as evaluators in multilingual scenarios. Figure \ref{fig:title_figure} shows an outline of our framework. The \systemname\ framework is an end-to-end pipeline, that starts with first creating a rich meta-evaluation dataset that contains a variety of samples across the metrics of interest. We do this by systematically prompting GPT-4 to generate a wide range of sample data points, that are then evaluated by native speakers. In the next step, we compare LLM judgments with human judgments. For this, we draw on our previous work \cite{hada-etal-2024-large} to prompt LLMs for evaluations and subsequently compare the scores with human judgments. In particular, for the task of \textbf{summarization} we create a dataset of 1000 summaries covering 10 languages, with human ratings across 5 metrics.\footnote{\systemname\ dataset and code available at \url{https://aka.ms/METAL}} Next, we obtain LLM evaluations from GPT-3.5-Turbo, GPT-4, and PaLM2 across these 1000 summaries and 5 metrics. 

Our findings show that the GPT-3.5-Turbo-based evaluator does not perform well across languages and metrics, while evaluators based on GPT-4 and PaLM2 perform better. We find that the evaluation ability of LLMs significantly across languages, motivating the creation of a meticulously crafted meta-evaluation dataset covering all target languages before using LLM-based evaluators. Lastly, our qualitative analysis shows that while GPT-4 and PaLM2 can achieve accuracy close to humans, the reasoning behind their evaluations is often flawed. While we study the applicability of the \systemname\ framework for the task of summarization, it is extensible to other tasks as well, by creating meta-evaluation datasets for the other tasks.

%% file: content/related_works.tex
\section{Related Work}
\label{sec:related_work}

\paragraph{Human Evaluation} Studies by \citet{kryscinski-etal-2018-improving,huang-etal-2020-achieved,shen-etal-2022-mred}  implemented the Likert scale for assessing various dimensions of generated summaries. \citet{fan-etal-2018-hierarchical,fabbri-etal-2019-multi,shen-etal-2022-sentbs} perform side-by-side comparisons of summaries produced by different models, using systems such as Elo for ranking the models based on performance.

\paragraph{Evaluation Datasets} Human-verified gold-standard datasets are crucial for being able to evaluate LLMs. \citet{skopek-etal-2023-towards} release \textit{riSum}, an English-centric dataset of document-instruction-output triplets, with the LLMs generating instructions and outputs, and a human evaluation is adopted to score the triplets, with a focus on ``instruction-following''. In our work, we manually curate the instructions and create a dataset covering 10 languages. \textsc{Seahorse} \cite{clark-etal-2023-seahorse} is a multilingual and multifaceted data for summarization with 96K summaries, and metrics related to grammar and output quality. They fine-tune T5 \cite{raffel2023exploring,xue-etal-2021-mt5} and PaLM \cite{chowdhery2022palm} models on the train split of the dataset for generating a spectrum of outputs, whereas we work with black-box models and tune our prompts to generate ``good'' and ``bad'' summaries.  

\paragraph{LLM Evaluation} Several previous studies have analyzed and evaluated LLMs on new tasks and standard benchmarks \cite{ahuja-etal-2023-mega, ahuja2023megaverse}. \citet{wang-etal-2023-chatgpt} prompt ChatGPT for summarization and other higher-level tasks, across various metrics and find a high correlation with human scores, with the caveat that it may be influenced by the way the meta-evaluation datasets are created. We extend this idea in the \systemname\ dataset. Other studies \cite{naismith-etal-2023-automated,fu2023gptscore,liu-etal-2023-g,zhang2023gpt4vision} have also put forth GPT scoring frameworks and prompt-based evaluators, however, they are mostly confined to English or Latin script languages. \citet{wu2023style} employ a Multi-Elo Rating System and advice for a similar multi-dimensional assessment of LLM-generated summaries. Previous studies \cite{li2023bactrianx,zheng2023judging,hada-etal-2024-large} have solely relied on GPT-4 as an LLM judge/evaluator. \citet{kim2023prometheus} propose a fine-tuned LLM, comparable to GPT-4 for input-evaluation rubric-output triplets, however, it is only fine-tuned for English. Previously, we performed a multilingual study of LLM meta-evaluation using an internal dataset of human judgments \citet{hada-etal-2024-large}. The dataset used in this work was not curated for the specific purpose of LLM-evaluator calibration and hence suffers from weaknesses such as dataset skew, and we focus this work on building a better dataset for meta-evaluation. We use the prompting strategies from our prior work \cite{hada-etal-2024-large} to evaluate our new curated dataset. To the best of our knowledge, no other study has proposed an end-to-end pipeline from generating a rich evaluation set to assessing the performance of LLM as evaluators in the multilingual scenario.

%% file: content/data.tex
\section{The \systemname\ Dataset}
The \systemname\ dataset contains a total of 1000 summaries across 10  languages. The dataset is specially curated to investigate the capabilities of LLMs as evaluators in different languages along 5 dimensions. In this section, we describe how the dataset was created and annotated.

\subsection{Dataset Creation}
\label{sect:dataset_creation}
The dataset consists of 100 summaries each, for 10 languages: English (En), French (Fr), Chinese Simplified (Zh), Hindi (Hi), Arabic (Ar), Bengali (Bn), Russian (Ru), Turkish (Tr), Japanese (Ja), and Swahili (Sw). We selected the languages to cover a diverse range of scripts and regions. The main text for each summary in our dataset was chosen from XL-Sum \cite{hasan-etal-2021-xl}, and the corresponding summary was generated by prompting GPT-4. A brief overview of our methodology is shown in Figure \ref{fig:title_figure}.

\paragraph{Main Text Selection} For each of the 10 languages we create a histogram of 20 bins of the number of tokens in the main text for all the datapoints in the test set of the XL-Sum dataset \cite{hasan-etal-2021-xl}. We chose 100 random samples from the bin with the highest frequency.

\paragraph{Summary Generation} To investigate the capabilities of LLMs as evaluators, our objective was to create an evaluation set of summaries with varying quality. To this end, for each of the chosen 1000 samples from the above step, we generate two summaries by prompting GPT-4 as follows. 

To generate good-quality summaries we provide the main text to GPT-4 and prompt it to return a summary of the main text such that it captures the essence of the main passage. We specifically ask for a summary that is highly rated on the 5 metrics of interest, described in the next section. We keep the temperature at 0 for the generation of good-quality summaries. 

To generate bad-quality summaries we provide the main text to GPT-4 and prompt it to act as an adversarial NLP assistant, and badly summarize the main passage.  We specifically ask for a summary that is rated low on the 5 metrics of interest. To generate bad quality summaries we keep the temperature as 1. In our initial experiments with lower temperatures, we observed that GPT-4 does not produce bad summaries even when specifically prompted to do so.

To further ensure the quality of summaries, in both styles of prompting we ask GPT-4 to also justify why the generated summary is good or bad.
Once we have 2 summaries per data point, we choose a good-quality summary or a bad-quality summary at random. The verbatim of our prompts are provided in \S\ref{sec:schema}

\subsection{Dataset Annotation}
\label{sec:data_annot}
For the 1000 summaries selected from the above process, we have each sample annotated by 3 annotators for 5 different metrics. We use the metrics described by \citet{hada-etal-2024-large} in their work: 

\paragraph{Linguistic Acceptability (LA)} This metric assesses whether the summary is acceptable to a native speaker. Specifically, the annotators are asked to determine whether the text exhibits signs of being translated, misuses words, or includes expressions that are not idiomatic in their language.

\paragraph{Output Content Quality (OCQ)} This metric assesses whether the general quality of the output text is good. The annotators are asked to consider flaws such as significant repetition, non-native language elements, or indications that the text has been web-scraped.

\paragraph{Task Quality (TQ)} This metric assesses the effectiveness of the summarization. It focuses on assessing the degree to which the summary aligns with key information in the main passage.

\paragraph{Problematic Content (PC)} This metric assesses the summary for the presence of any content that may be deemed offensive, inappropriate, or harmful. It serves as a filter against outputs that might perpetuate harmful stereotypes or misinformation. 

\paragraph{Hallucinations (H)} This metric assesses whether the summary remains anchored to, and consistent with, the main passage. It serves as a check against unwarranted deviations from the ground truth provided in the input. \\ 

\noindent For LA, OCQ, and TQ, annotators were asked to assign one of the three possible classes: Bad (0), Medium (1), Good (2). For PC and H, annotators were asked to assign one of the two possible classes: Present (1) and Absent (0).\footnote{We include the detailed annotation instructions in Appendix \S\ref{sec:task_ins}}

\paragraph{Annotation Task and Quality} Each datapoint was annotated by three annotators for the five metrics. Annotators were native speakers of the respective language and trained professionals contracted through an external annotator services company. The pay was adjusted based on the annotator's region and experience. Since we wanted to ensure we had a strong evaluation set to study the capabilities of LLMs as evaluators, special attention was given to the quality of annotations. The annotators were specifically trained to perform annotations for this task and a sample of annotations was reviewed for all annotators. Annotations were reviewed for accuracy and guideline consistency. Based on the review, feedback was provided to the annotators, and ambiguous cases were re-annotated.

\begin{figure}[t!]
    \centering
    \includegraphics[width=0.8\columnwidth]{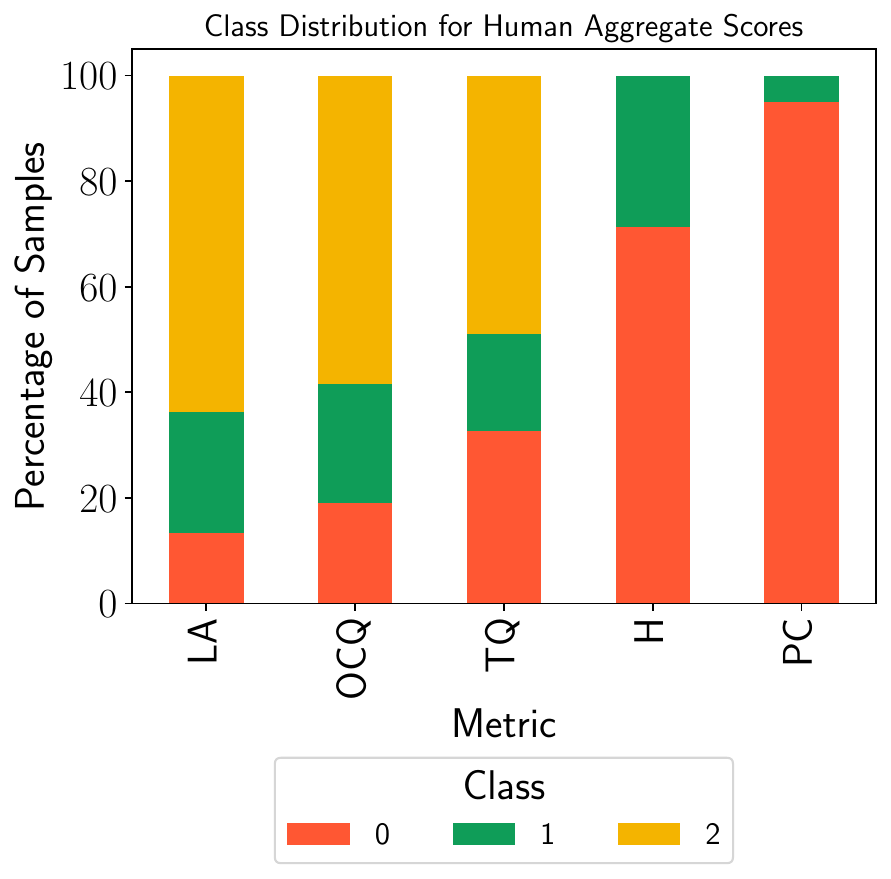}
    \caption{Class distribution for various metrics, summed over all languages.}
    \label{fig:class_dist_humans}
\end{figure}

Table \ref{tab:agreement} in Appendix \S\ref{sec:agreement} shows the Fleiss' Kappa ($\kappa$) and pairwise agreement (computed as F1)  values among the annotators for the various languages and metrics. All our $\kappa$ values are $> 0.6$ (except for H in En, $\kappa = 0.54$), and all F1 values are $> 0.75$, indicating substantial agreement. Some $\kappa$ values are 0 due to class skew, but high F1 in these cases indicates high reliability. For our experiments, we take the majority vote from the three human annotations per sample as the aggregate class for that sample. In the case of 3 distinct annotations, we take the average value. Figure \ref{fig:class_dist_humans} shows the distribution of the aggregate annotations over various languages and metrics. 

\subsection{Dataset Statistics}
\label{sec:dataset_stat}
As discussed in \S\ref{sect:dataset_creation}, we sample the datapoints from XL-Sum based on the number of tokens of the passage. Specifically, the \texttt{tiktoken}\footnote{\url{https://github.com/openai/tiktoken}} was utilized for the tokenization process, and the length (token) distribution of the passages and summaries are presented in Table \ref{tab:combined} along with the number of good and bad instances per language. Table \ref{tab:good_bad_class_dist} presents the frequency/distribution of the classes (0, 1, and 2) in the good and bad summaries. Notably, the first row of the table depicts higher counts of low scores (class 0) for the bad summaries, relative to the good ones. Further, the medium scores (class 1) also contain a higher frequency in the bad summaries, however, the difference between the frequencies is lower than that of class 0. Surprisingly, the bad summaries have more of class 2 scores than the good ones (third row) in Linguistic Acceptability. This goes on to show that LLMs are incapable of generating incoherent text despite adversarial prompts.

\begin{table}[t!]
\tiny
\centering 
\begin{tabular}{@{}lcccccc@{}}
\toprule
\textbf{Lang} & \textbf{Passage} & \textbf{Summary} & \textbf{Good} & \textbf{Bad} \\ \midrule
AR & 877.39 $\pm$ 53.00 & 160.70 $\pm$ 87.29 & 50 & 50 \\
BN & 4161.58 $\pm$ 534.91 & 339.83 $\pm$ 160.55 & \textbf{53} & 47 \\
EN & 358.29 $\pm$ 21.09 & 67.71 $\pm$ 29.57 & 46 & \textbf{54} \\
FR & 341.96 $\pm$ 26.89 & 84.79 $\pm$ 39.27 & \textbf{51} & 49 \\
HI & 1234.82 $\pm$ 70.28 & 219.08 $\pm$ 92.38 & 48 & \textbf{52} \\
JA & 1327.44 $\pm$ 61.50 & 136.44 $\pm$ 81.11 & \textbf{52} & 48 \\
RU & 748.26 $\pm$ 47.52 & 139.09 $\pm$ 72.28 & 43 & \textbf{57} \\
SW & 518.70 $\pm$ 35.90 & 127.98 $\pm$ 73.79 & 47 & \textbf{53} \\
TR & 625.77 $\pm$ 40.96 & 136.44 $\pm$ 68.76 & 42 & \textbf{58} \\
ZH & 666.03 $\pm$ 47.78 & 124.16 $\pm$ 67.80 & 48 & \textbf{52} \\  \bottomrule
\end{tabular}
\caption{Length distribution and number of instances per language}
\label{tab:combined}
\end{table}

\begin{table}[t!]
\tiny
\centering
\begin{tabular}{@{}lccccc@{}}
\toprule
\textbf{Class} & \textbf{LA} & \textbf{OCQ} & \textbf{TQ} & \textbf{H} & \textbf{PC} \\ \midrule
0 & 54 / \textbf{80} & 78 / \textbf{112} & 124 / \textbf{202} & 352 / \textbf{362} & 457 / \textbf{493} \\
1 & 113 / \textbf{116} & 104 / \textbf{121} & 91 / \textbf{93} & 128 / \textbf{158} & 23 / \textbf{27} \\
2 & 313 / \textbf{324} & \textbf{298} / 287 & \textbf{265} / 225 & - / - & - / - \\ \bottomrule
\end{tabular}
\caption{Class distribution for various metrics, N(Good)/N(Bad). The highest frequency is bolded.}
\label{tab:good_bad_class_dist}
\end{table}

%% file: content/experiments.tex
\section{Experiments}

\subsection{Models}
GPT-4-32K \cite{openai2023gpt4}, GPT-3.5-Turbo \cite{ouyang2022training}, and PaLM2 Text-Bison \cite{anil2023palm} models were used as the evaluators to score the LLM-generated summary according to the given metrics. \footnote{Both GPT models were accessed through Azure and PaLM2 via VertexAI.}

\subsection{Prompts}
The models were prompted using the LangChain\footnote{\url{https://github.com/langchain-ai/langchain}} framework and a structured JSON output format was maintained to parse the generations efficiently. The prompts for evaluation follow the same verbatim as \citet{hada-etal-2024-large}.

\subsection{Prompting Strategies}
Based on our previous work \cite{hada-etal-2024-large}, we use the simple and detailed prompting strategies for all models, and all the metrics are evaluated independently in a single call to the API. All prompts were provided in English, as \citet{ahuja-etal-2023-mega} have shown that multilingual instructions lead to worse performance. Further, the temperature is set to $0$ for reproducibility.

\paragraph{Simple Instruction} A rudimentary description of the metric and scoring schema is provided, as shown in Figure \ref{fig:metricdescription_LA} in appendix \S\ref{sec:simple_instructions}.

\paragraph{Detailed Instruction} An informative and thorough description of the metric and a case-by-case breakdown of the scoring schema is provided, as shown in Figure \ref{fig:metricdescription_complex_LA}in the appendix \S\ref{sec:simple_instructions}. 

\subsection{Meta Evaluation}
\label{sec:calibration}
As described in \S\ref{sec:data_annot} we use the aggregate of the three annotations for our experiments. 

\paragraph{Pairwise Agreement (F1)} We measure the pairwise agreement between the LLM evaluators and human aggregate scores per language and metric. To account for any class imbalance, we report the weighted F1 score instead of accuracy. 

\paragraph{Class Distribution} We analyze the class distribution of the human aggregate scores and the various model predictions for three possible classes: When all three annotators agree, when two of three annotators agree, and when no annotators agree. We do this analysis only for metrics with 3 possible classes: LA, OCQ, and TQ. 

\subsection{Comparison between \textsc{Seahorse} and \systemname} \textsc{Seahorse} \cite{clark-etal-2023-seahorse} is a dataset akin to \systemname, as described in \S\ref{sec:related_work}. It contains summaries generated using several models for passages from popular summarization datasets such as XL-Sum \cite{hasan-etal-2021-xl}, XSum \cite{narayan-etal-2018-dont}, MLSum \cite{scialom-etal-2020-mlsum} and WikiLingua \cite{ladhak-etal-2020-wikilingua}. We use the XL-Sum subset of Seahorse and find out common datapoints between \textsc{Seahorse} and \systemname. There are a total of 27 overlapping data points: 1 in English, 10 in Russian, and 16 in Turkish. These datapoints can have one or more summaries in \textsc{Seahorse} generated by \texttt{mt5\_small}: The 300M version of
mT5 \cite{xue-etal-2021-mt5}, \texttt{mt5\_small\_250}: The same \texttt{mt5\_small} model
but using the checkpoint after training 250 steps, \texttt{mt5\_xxl}: The 13B mT5 model, \texttt{palm\_1shot}: 540B PaLM
model \cite{chowdhery2022palm} prompted
with one in-domain example, \texttt{palm\_finetuned}: 540B PaLM model finetuned on
training data for the respective dataset.
We use our detailed prompting strategy to evaluate the summaries generated by various models in \textsc{Seahorse} for our metrics and compare them with the evaluation of the summaries generated by GPT-4 for the same main passages in \systemname.\ We use PaLM2 and GPT-4 as evaluators. \footnote{We do not consider the 1 overlapping datapoint in English for our experiment.} 

%% file: content/results_discussions.tex
\section{Results and Discussions}
\label{sec:results}

\subsection{Pairwise Agreement (F1)}

\begin{table*}[t!]
\tiny
\centering
\begin{tabular}{@{}lllcccccccccc@{}}
\toprule
\textit{\textbf{Metric}} & \textit{\textbf{\begin{tabular}[c]{@{}l@{}}Prompting\\ Strategy\end{tabular}}} & \textit{\textbf{Model}} & \textbf{AR} & \textbf{BN} & \textbf{EN} & \textbf{FR} & \textbf{HI} & \textbf{JA} & \textbf{RU} & \textbf{SW} & \textbf{TR} & \textbf{ZH} \\ \midrule
\multirow{7}{*}{\textit{\textbf{LA}}} & \textit{\textbf{}} & \textit{\textbf{human}} & 0.89 & 0.81 & 0.86 & 0.99 & 0.87 & 0.95 & 0.98 & 0.82 & 0.97 & 0.94 \\ \cmidrule(l){2-13} 
 & \multirow{3}{*}{\textit{\textbf{Simple}}} & \textit{\textbf{GPT-3.5-Turbo}} & 0.54 & 0.43 & 0.61 & 0.61 & 0.44 & 0.45 & 0.67 & 0.78 & 0.55 & 0.59 \\
 &  & \textit{\textbf{GPT-4}} & 0.74 & 0.15 & 0.72 & 0.88 & 0.59 & 0.48 & 0.74 & 0.61 & 0.72 & 0.85 \\
 &  & \textit{\textbf{PaLM2}} & 0.74 & 0.11 & 0.54 & 0.73 & 0.64 & 0.38 & 0.77 & 0.82 & 0.69 & 0.84 \\ \cmidrule(l){2-13} 
 & \multirow{3}{*}{\textit{\textbf{Detailed}}} & \textit{\textbf{GPT-3.5-Turbo}} & 0.19 & 0.44 & 0.59 & 0.40 & 0.18 & 0.19 & 0.53 & 0.57 & 0.15 & 0.19 \\
 &  & \textit{\textbf{GPT-4}} & 0.71 & 0.22 & 0.82 & 0.81 & 0.61 & 0.47 & 0.80 & 0.76 & 0.72 & 0.85 \\
 &  & \textit{\textbf{PaLM2}} & 0.71 & 0.21 & 0.54 & 0.75 & 0.59 & 0.34 & 0.78 & 0.88 & 0.64 & 0.84 \\ \midrule
\multirow{7}{*}{\textit{\textbf{OCQ}}} & \textit{\textbf{}} & \textit{\textbf{human}} & 0.85 & 0.82 & 0.82 & 0.97 & 0.83 & 0.93 & 0.93 & 0.84 & 0.84 & 0.91 \\ \cmidrule(l){2-13} 
 & \multirow{3}{*}{\textit{\textbf{Simple}}} & \textit{\textbf{GPT-3.5-Turbo}} & 0.11 & 0.39 & 0.65 & 0.47 & 0.17 & 0.21 & 0.64 & 0.61 & 0.52 & 0.33 \\
 &  & \textit{\textbf{GPT-4}} & 0.71 & 0.27 & 0.69 & 0.70 & 0.65 & 0.47 & 0.94 & 0.85 & 0.69 & 0.88 \\
 &  & \textit{\textbf{PaLM2}} & 0.69 & 0.23 & 0.63 & 0.68 & 0.58 & 0.43 & 0.92 & 0.91 & 0.67 & 0.79 \\ \cmidrule(l){2-13} 
 & \multirow{3}{*}{\textit{\textbf{Detailed}}} & \textit{\textbf{GPT-3.5-Turbo}} & 0.23 & 0.54 & 0.59 & 0.50 & 0.31 & 0.33 & 0.64 & 0.58 & 0.50 & 0.44 \\
 &  & \textit{\textbf{GPT-4}} & 0.69 & 0.26 & 0.68 & 0.72 & 0.65 & 0.51 & 0.92 & 0.88 & 0.68 & 0.84 \\
 &  & \textit{\textbf{PaLM2}} & 0.68 & 0.29 & 0.57 & 0.65 & 0.66 & 0.41 & 0.92 & 0.91 & 0.69 & 0.86 \\ \midrule
\multirow{7}{*}{\textit{\textbf{TQ}}} & \textit{\textbf{}} & \textit{\textbf{human}} & 0.77 & 0.78 & 0.77 & 0.90 & 0.78 & 0.99 & 0.94 & 0.84 & 0.87 & 0.82 \\ \cmidrule(l){2-13} 
 & \multirow{3}{*}{\textit{\textbf{Simple}}} & \textit{\textbf{GPT-3.5-Turbo}} & 0.63 & 0.53 & 0.52 & 0.84 & 0.58 & 0.81 & 0.83 & 0.82 & 0.65 & 0.77 \\
 &  & \textit{\textbf{GPT-4}} & 0.60 & 0.64 & 0.53 & 0.81 & 0.56 & 0.87 & 0.95 & 0.87 & 0.60 & 0.78 \\
 &  & \textit{\textbf{PaLM2}} & 0.56 & 0.67 & 0.41 & 0.83 & 0.56 & 0.85 & 0.90 & 0.88 & 0.59 & 0.79 \\ \cmidrule(l){2-13} 
 & \multirow{3}{*}{\textit{\textbf{Detailed}}} & \textit{\textbf{GPT-3.5-Turbo}} & 0.26 & 0.49 & 0.54 & 0.76 & 0.22 & 0.44 & 0.63 & 0.63 & 0.58 & 0.31 \\
 &  & \textit{\textbf{GPT-4}} & 0.71 & 0.64 & 0.59 & 0.86 & 0.66 & 0.86 & 0.96 & 0.87 & 0.63 & 0.76 \\
 &  & \textit{\textbf{PaLM2}} & 0.58 & 0.66 & 0.38 & 0.83 & 0.51 & 0.84 & 0.94 & 0.90 & 0.65 & 0.73 \\ \midrule \midrule
\multirow{7}{*}{\textit{\textbf{H}}} & \textit{\textbf{}} & \textit{\textbf{human}} & 0.89 & 0.97 & 0.85 & 0.97 & 0.90 & 0.99 & 0.99 & 0.93 & 0.84 & 1.00 \\ \cmidrule(l){2-13} 
 & \multirow{3}{*}{\textit{\textbf{Simple}}} & \textit{\textbf{GPT-3.5-Turbo}} & 0.54 & 0.27 & 0.81 & 0.75 & 0.36 & 0.63 & 0.72 & 0.66 & 0.57 & 0.59 \\
 &  & \textit{\textbf{GPT-4}} & 0.93 & 0.74 & 0.85 & 0.91 & 0.89 & 0.94 & 0.93 & 0.90 & 0.87 & 0.90 \\
 &  & \textit{\textbf{PaLM2}} & 0.94 & 0.77 & 0.78 & 0.92 & 0.90 & 0.82 & 0.72 & 0.80 & 0.76 & 0.87 \\ \cmidrule(l){2-13} 
 & \multirow{3}{*}{\textit{\textbf{Detailed}}} & \textit{\textbf{GPT-3.5-Turbo}} & 0.06 & 0.01 & 0.42 & 0.58 & 0.09 & 0.36 & 0.50 & 0.37 & 0.22 & 0.19 \\
 &  & \textit{\textbf{GPT-4}} & 0.95 & 0.72 & 0.85 & 0.90 & 0.88 & 0.96 & 0.94 & 0.89 & 0.86 & 0.88 \\
 &  & \textit{\textbf{PaLM2}} & 0.91 & 0.73 & 0.76 & 0.90 & 0.86 & 0.94 & 0.87 & 0.87 & 0.86 & 0.91 \\ \midrule
\multirow{7}{*}{\textit{\textbf{PC}}} & \textit{\textbf{}} & \textit{\textbf{human}} & 0.93 & 1.00 & 1.00 & 1.00 & 0.94 & 0.99 & 0.99 & 0.86 & 1.00 & 1.00 \\ \cmidrule(l){2-13} 
 & \multirow{3}{*}{\textit{\textbf{Simple}}} & \textit{\textbf{GPT-3.5-Turbo}} & 0.52 & 0.23 & 0.83 & 0.56 & 0.32 & 0.31 & 0.33 & 0.51 & 0.45 & 0.63 \\
 &  & \textit{\textbf{GPT-4}} & 0.90 & 0.99 & 1.00 & 0.95 & 0.85 & 1.00 & 0.97 & 0.73 & 1.00 & 0.97 \\
 &  & \textit{\textbf{PaLM2}} & 0.89 & 1.00 & 0.97 & 0.85 & 0.86 & 0.95 & 0.92 & 0.71 & 0.99 & 0.96 \\ \cmidrule(l){2-13} 
 & \multirow{3}{*}{\textit{\textbf{Detailed}}} & \textit{\textbf{GPT-3.5-Turbo}} & 0.28 & 0.06 & 0.68 & 0.45 & 0.23 & 0.28 & 0.20 & 0.43 & 0.28 & 0.36 \\
 &  & \textit{\textbf{GPT-4}} & 0.87 & 0.99 & 0.99 & 0.87 & 0.85 & 0.95 & 0.91 & 0.71 & 0.91 & 0.96 \\
 &  & \textit{\textbf{PaLM2}} & 0.89 & 0.84 & 0.97 & 0.88 & 0.86 & 0.79 & 0.88 & 0.80 & 0.92 & 0.95 \\ \bottomrule
\end{tabular}
\caption{F1 scores for various languages, models, and prompting strategies.}
\label{tab:F1-scores-table}
\end{table*}

Table \ref{tab:F1-scores-table} and Figures \ref{fig:spider-3-plots} and \ref{fig:spider-2-plots} in Appendix \S\ref{f1_scores_radar_plots} present the distribution of F1 scores of various models with the two prompting strategies on the 10 languages. For ``human scores'', we average the pairwise F1 scores of all the annotators, i.e., A1-A2, A2-A3, and A3-A1. For the ``model scores'' in the plot, the F1 score between the annotator aggregate and model evaluation is computed. 

For all the metrics, humans have the best agreement. In the case of LA, for most of the languages, GPT-4 with detailed instructions performs the closest to humans, followed by GPT-4 with simple instructions. GPT-3.5-Turbo performs the worst with detailed instructions and a significant difference is observed by making the instructions simple, however, no difference is found for English. For most of the languages, especially Zh, Hi, Ru, and Ar, GPT-4 and PaLM2 perform similarly. 

For OCQ, GPT-3.5-Turbo performs the worst, and detailed instruction improves the performance marginally over simple instructions. GPT-4 and PaLM2 perform very closely to humans for Russian, however, there is a gap between the human and LLM scores on the rest of the languages. For TQ, both prompting strategies for GPT-4 and PaLM2 do equally well on most languages except Ar, Hi, Zh, and En. In these cases, GPT-4 with detailed instructions does the best.

For PC and H, all models show very similar scores as compared to humans, except GPT-3.5-Turbo. Simple instructions for GPT-3.5-Turbo improve performance for both metrics with a higher gain on H. Interestingly, for Bn and metrics LA and OCQ, both prompting strategies for GPT-3.5-Turbo do better than GPT-4 and PaLM2. For Sw and metrics LA, OCQ, and TQ, the agreement between humans and GPT-4 or PaLM2 is as good as the agreement between humans. GPT-3.5-Turbo with detailed instructions does worse than GPT-3.5-Turbo for all metrics except OCQ.  

Overall, we find that the performance of GPT-4 and especially PaLM2 are largely independent of simple or detailed instructions, in all languages. The same holds for GPT-3.5-Turbo only on English, suggesting that it is less sensitive to prompting in English. GPT-4 with detailed instructions comes closest to human evaluation, with marginal improvements over simple instructions, in most cases. GPT-4 and PaLM2 are very effective in identifying hallucinations and problematic content for all languages. 

\subsection{Class Distribution}
\begin{figure}[t!]
    \centering
    \begin{subfigure}[t]{\columnwidth}
        \includegraphics[width=\textwidth]{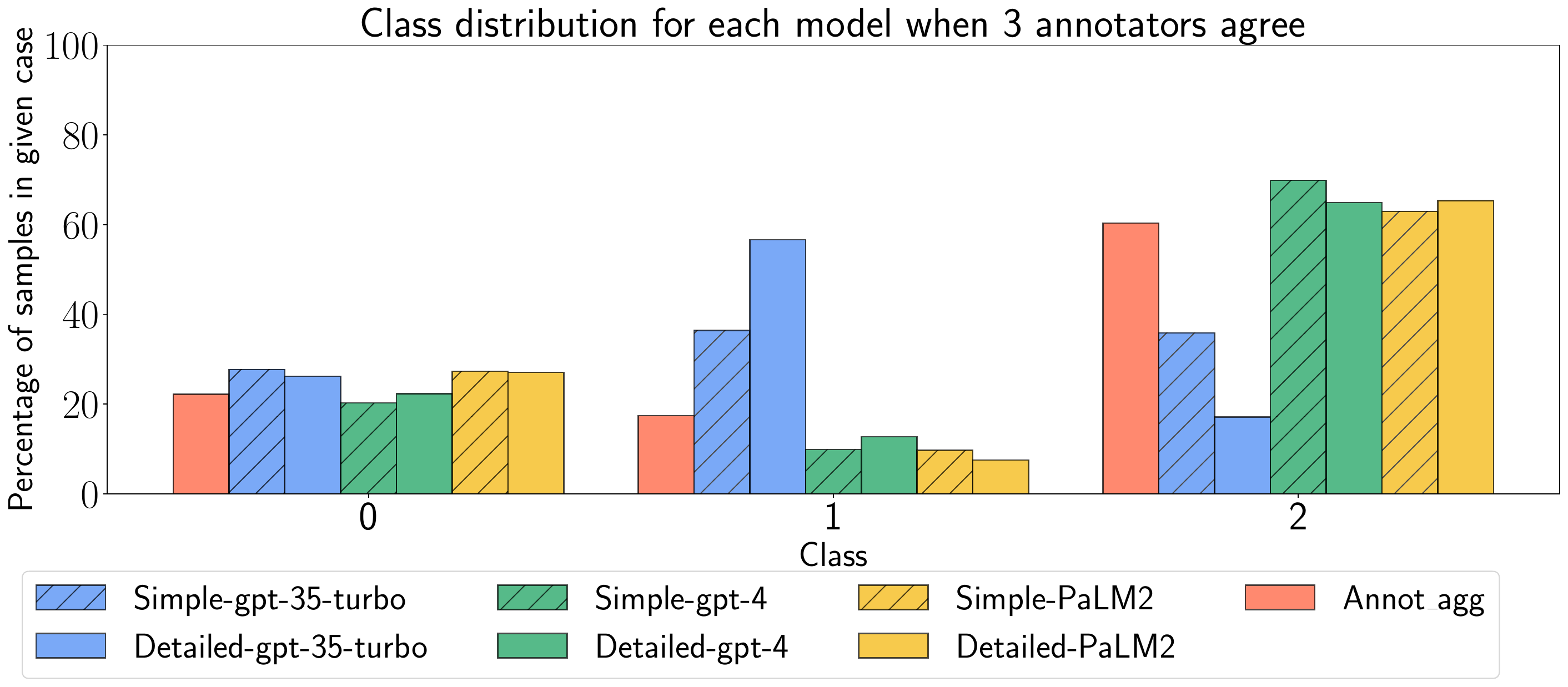}
        \caption{}
        \label{fig:three_annots}
    \end{subfigure}
    \\
    \begin{subfigure}[t]{\columnwidth}
        \includegraphics[width=\textwidth]{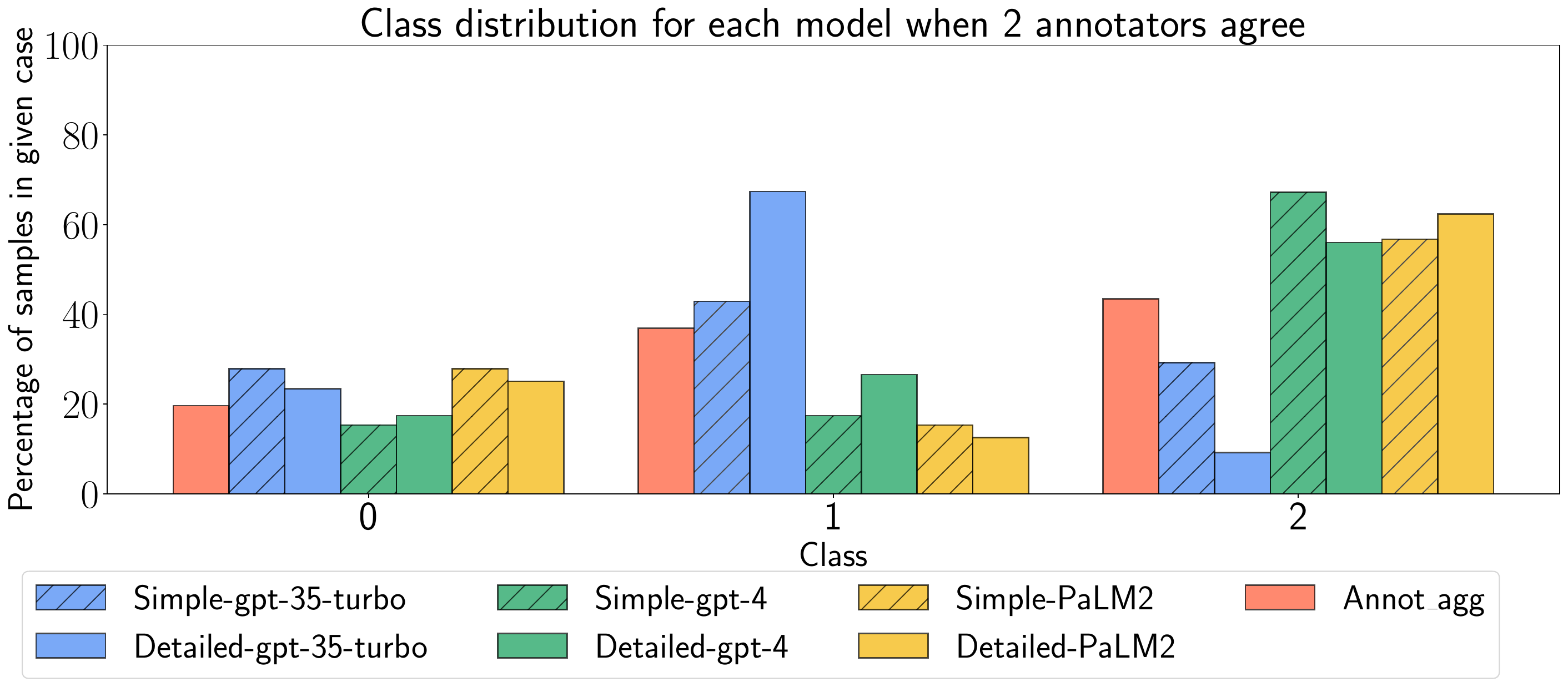}
        \caption{}
        \label{fig:two_annots}
    \end{subfigure}
    \\
    \begin{subfigure}[t]{\columnwidth}
        \includegraphics[width=\textwidth]{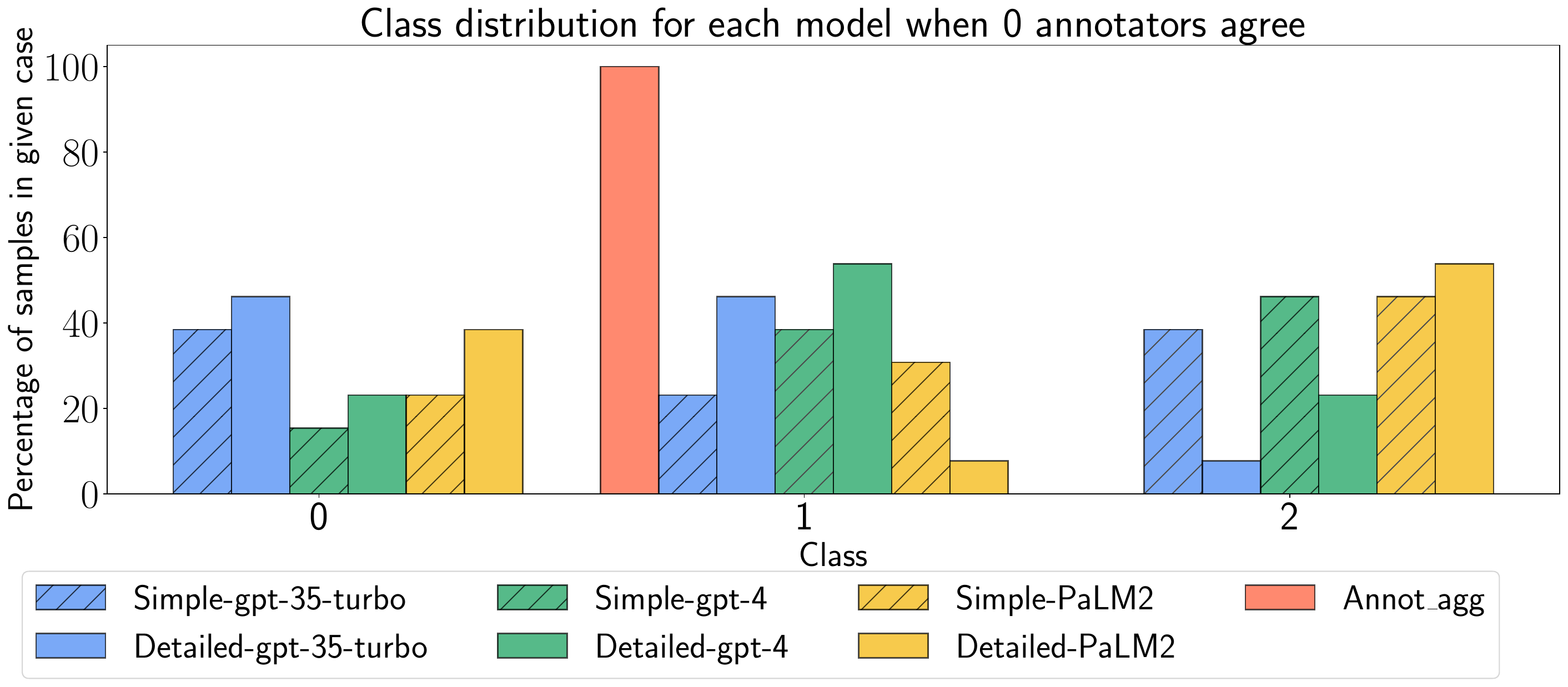}
        \caption{}
        \label{fig:no_annots}
    \end{subfigure}
    \caption{Class distribution for different cases}
    \label{fig:classdist}
\end{figure}

Figure \ref{fig:classdist} shows the distribution of human aggregate score and various models for the three cases. 
In the case where all annotators agree, as shown in Figure \ref{fig:three_annots} we can see that the class distribution for GPT-4 and PaLM2 with both prompting variations is very close to the class distribution of human aggregate scores. This indicates that when humans have full agreement (perhaps due to easier samples), LLM-based evaluators also perform well.

In the case where two of three annotators agree, we can see in Figure \ref{fig:two_annots} that for both prompting variations GPT-4 and PaLM2 often over-predict class 2, under-predict class 1 and are similar to humans for class 0. Overall, detailed-GPT-4 comes closest to the distribution of human aggregate scores. For both cases, GPT-3.5-Turbo often prefers the middle class, which can be indicative of scale region bias. 

In our third case, owing to our high quality of annotations, we have only 13 out of 3000 samples where no annotators agree.  Figure \ref{fig:no_annots} shows the class distribution for this case. We can observe that GPT-3.5-Turbo with simple instructions assigns different classes with almost equal frequency. GPT-4 with detailed instructions often outputs the middle class, which is the annotator aggregate as well. PaLM2 with detailed instructions outputs the highest or the lowest score, and interestingly very few times opts for the middle class.

From this analysis, we conclude that while simple or detailed instructions for both GPT-4 and PaLM2 perform equally well when all human annotators agree, detailed instructions for GPT-4 do best when there is disagreement amongst annotators.

\subsection{Comparison of evaluation between \textsc{Seahorse} and \systemname}

\begin{table*}[t!]
\tiny
\centering
\resizebox{\textwidth}{!}
{
\begin{tabular}{@{}llcc|cc|cc|cc|cc|cc|cc|cc@{}}
\toprule
\textbf{Lang} &
  \textbf{Metric} &
  \multicolumn{10}{c|}{\textbf{\textsc{Seahorse}}} &
  \multicolumn{2}{c|}{\textbf{XL-Sum}} &
  \multicolumn{4}{c}{\textbf{\systemname}} \\ \midrule
 &
   &
  \multicolumn{2}{c}{\texttt{$mt5\_small\_250$}} &
  \multicolumn{2}{c}{\texttt{$mt5\_xxl$}} &
  \multicolumn{2}{c}{\texttt{$mt5\_small$}} &
  \multicolumn{2}{c}{\texttt{$palm\_1shot$}} &
  \multicolumn{2}{c|}{\texttt{$palm\_finetuned$}} &
  \multicolumn{2}{c|}{\texttt{reference}} &
  \multicolumn{2}{c}{\texttt{$GPT-4_{good}$}} &
  \multicolumn{2}{c}{\texttt{$GPT-4_{bad}$}} \\ \midrule
  &
   &
  PaLM2 &
  GPT-4 &
  PaLM2 &
  GPT-4 &
  PaLM2 &
  GPT-4 &
  PaLM2 &
  GPT-4 &
  PaLM2 &
  GPT-4 &
  PaLM2 &
  GPT-4 &
  PaLM2 &
  GPT-4 &
  PaLM2 &
  GPT-4 \\ \midrule
\multirow{5}{*}{\textbf{\textit{RU}}} &
  \textbf{\textit{H}} &
  0.60 &
  0.40 &
  0.00 &
  0.20 &
  0.43 &
  0.43 &
  0.20 &
  0.00 &
  0.00 &
  0.00 &
  0.00 &
  0.00 &
  1.00 &
  1.00 &
  0.75 &
  0.75 \\
 &
  \textbf{\textit{LA}} &
  0.60 &
  1.00 &
  1.40 &
  2.00 &
  0.57 &
  2.00 &
  1.60 &
  2.00 &
  1.25 &
  2.00 &
  1.33 &
  2.00 &
  0.00 &
  1.00 &
  0.62 &
  0.87 \\
 &
  \textbf{\textit{OCQ}} &
  0.60 &
  0.60 &
  1.40 &
  1.20 &
  0.57 &
  0.86 &
  1.40 &
  1.60 &
  1.00 &
  1.25 &
  1.00 &
  1.67 &
  0.0 &
  0.00 &
  0.50 &
  0.50 \\
 &
  \textbf{\textit{TQ}} &
  0.60 &
  0.40 &
  1.00 &
  1.20 &
  0.71 &
  0.71 &
  1.60 &
  1.20 &
  1.00 &
  1.00 &
  1.00 &
  1.00 &
  0.00 &
  0.00 &
  0.50 &
  0.50 \\
 &
  \textbf{\textit{PC}} &
  0.00 &
  0.00 &
  0.00 &
  0.00 &
  0.14 &
  0.00 &
  0.00 &
  0.00 &
  0.00 &
  0.00 &
  0.00 &
  0.00 &
  0.00 &
  0.00 &
  0.25 &
  0.25 \\ \midrule
\multirow{5}{*}{\textbf{\textit{TR}}} &
  \textbf{\textit{H}} &
  0.55 &
  0.75 &
  0.10 &
  0.10 &
  0.50 &
  0.50 &
  0.00 &
  0.12 &
  0.00 &
  0.00 &
  0.00 &
  0.00 &
  0.14 &
  0.28 &
  0.55 &
  0.55 \\
 &
  \textbf{\textit{LA}} &
  0.11 &
  1.12 &
  1.40 &
  2.00 &
  0.75 &
  1.62 &
  1.50 &
  2.00 &
  1.20 &
  2.00 &
  1.33 &
  2.00 &
  1.57 &
  1.71 &
  0.78 &
  1.55 \\
 &
  \textbf{\textit{OCQ}} &
  0.11 &
  0.62 &
  1.20 &
  1.50 &
  0.62 &
  1.00 &
  1.37 &
  1.62 &
  1.00 &
  1.60 &
  1.33 &
  1.67 &
  1.57 &
  1.57 &
  0.78 &
  1.00 \\
 &
  \textbf{\textit{TQ}} &
  0.11 &
  0.50 &
  1.20 &
  1.30 &
  0.50 &
  0.75 &
  1.37 &
  1.25 &
  0.80 &
  1.20 &
  1.17 &
  1.33 &
  1.57 &
  1.57 &
  0.78 &
  1.11 \\
 &
  \textbf{\textit{PC}} &
  0.33 &
  0.00 &
  0.00 &
  0.00 &
  0.00 &
  0.12 &
  0.00 &
  0.00 &
  0.00 &
  0.00 &
  0.00 &
  0.00 &
  0.00 &
  0.00 &
  0.33 &
  0.22 \\ \bottomrule
\end{tabular}%
}
\caption{Evaluation of overlapping summaries generated by various models in \textsc{Seahorse} and GPT-4 in \systemname\ for RU and TR}
\label{tab:seahorse_comp}
\end{table*}

Table \ref{tab:seahorse_comp} shows the values for evaluation of overlapping datapoints between \textsc{Seahorse} and \systemname\ for RU and TR. The values are averaged over all datapoints. For Russian, we can observe that \texttt{palm\_1shot} does the best as it is rated highly across all metrics by both the models. Interestingly, \texttt{palm\_1shot} is rated better than \texttt{palm\_finetuned}.  GPT-4$_{good}$ summaries received very bad evaluations across all metrics by both models. On further investigation, we found that GPT-4$_{good}$ category had only 2 datapoints out of the overlapping 10 datapoints, and co-incidentally these 2 generated summaries were of bad quality and have been rated poorly by human annotators as well. This indicates that GPT-4 might not always be biased towards its own generations as compared to generations from other models. We can also observe that in almost all cases GPT-4 provides a higher rating as compared to PaLM2. For Turkish, we can observe that GPT-4$_{good}$ receives the highest ratings by both models. Similar to Russian, \texttt{palm\_1shot} receives better ratings than palm\_finetuned. As expected, \texttt{mt5\_small\_250} receives the lowest ratings since it is an under-trained model. We can also notice a clear difference in ratings for GPT-4$_{good}$ vs GPT-4$_{bad}$ generations. Overall, from this experiment we can conclude that using our metrics and prompting methods we can compare generations from different models.

%% file: content/qualitative_analysis.tex
\section{Qualitative Analysis}

\begin{table*}[t!]
{\tiny
\begin{tabularx}{\textwidth}{lllXXX}
\toprule
\textbf{No.}&
\textbf{Language} &
  \textbf{Metric} &
  \textbf{Human Score and Comments} &
  \textbf{GPT-4 Score and Justification} &
  \textbf{PaLM2 Score and Justification} \\ \midrule
1 & English &
  TQ &
  `A1': 1, `Some grammatical errors present. Some misaligned elements and omission of important information.', `A2': 2, `High quality summary, except   missing verbs so medium linguistic acceptability',  `A3': 0, `Summary contains grammatical errors, has poor discourse and is incoherent.' &
  2, The summary adheres to the instructions and captures the main points of the passage. Therefore,  the summary demonstrates a precise understanding and adherence to the prompt's instructions. &
  2, The summary captures the main points of the passage. It correctly identifies the key players in the dispute, the reasons for the dispute, and the potential consequences of the dispute. The summary is concise and well-written. \\ \midrule
2 & Arabic &
  LA &
  `A1': 1, `Text seems to be copied. Repetition noticed. Summary is too   short more like a headline.', `A2': 2, `it is missing a lot of content from the   main passage ',  `A3': 1, 'The summary is very brief and does not meet all the   points mentioned in the main passage.' &
  2, The vocabulary used is contextually relevant and the sentences reflect the natural rhythm, emphasis, and intonation of spoken Arabic. it does not violate any linguistic norms or rules, and thus scores high on linguistic acceptability. &
  2, The summary is grammatically correct and uses contextually relevant vocabulary. However, it is overly verbose and lacks natural rhythm and intonation. \\
3 & French &
  OCQ &
  `A1': 2, `Le résumé, bien que mal conçu, affirme qu'une femme accuse  Bill Cosby d'agression sexuelle à cause de l'argent, ce qui n'est pas vrai    selon le passage original.  Il y a des   hallucinations dans le résumé et la qualité de la tâche est mauvaise.",   `A2': 2, `Le résumé fourni contient des informations précises sur l’affaire  Bill   Cosby, telles que les allégations contre lui et le règlement  civil avec l’un   des accusateurs.', `A3': 1, `Insted of 'veut' it would have been better to use ``ne voudrait'' or ``est accusée de ne vouloir que de l'argent".  To me, it is a fake news.' &
  0, The summary provided does not accurately reflect the content of the passage. It also inaccurately portrays the accuser as someone who just wants money, without mentioning her testimony or the fact that Cosby settled the case. &
  0, The summary does not capture the essence of the passage. It is not clear, coherent, or engaging. It does not convey the main points of the passage and it is not original. \\ \midrule
4 & Hindi &
  TQ &
  `A1': 2, `The summary is counterfactual in relation to the input. The president condemned the incident and in summary it states that he praised the incident.  This is a problematic content because it involves student violence,   gun violence, teen violence, etc.… less or no grammatical errors makes the   task accuracy good.', `A2': 2, `The summary does not justifies the actual sentence.  The quality of the summary is very bad.',  `A3': `2, no comments' &
  0, The summary provided does not adhere to the passage at all. It misrepresents the events described in the passage. &
  0, The summary does not adhere to the instructions given in the prompt. The prompt asks for a summary of the passage, but the summary is not a summary of the passage. The summary is a   made-up story that is not related to the passage. \\
5 & \begin{tabular}[c]{@{}l@{}}Chinese\\ Simplified\end{tabular} &
  OCQ &
  `A1': 2, `Good summary without mistakes', `A2': 2, `The summary fails to mention the important message of the original, which is, many civilians were killed. The summary only indicates that some people were killed,  `A3': 2, `Key information is missing in the summarization. The summarization is   too general.' &
  1, The summary is grammatically correct and uses appropriate language. However, it lacks clarity and relevance to the passage. &
  1, The summary is concise and captures the main points of the passage. It is also well-written and easy to understand.  However, it does not provide any new information or insights beyond what is already in the passage. \\ \bottomrule
\end{tabularx}}
\caption{Samples from \systemname dataset with human and LLM (detailed variation) scores and comments. Note: We only include snippets from the full comment.}
\label{tab:QT}
\end{table*}

While results in section \S\ref{sec:results} show that detailed instructions to GPT-4 and PaLM2 give evaluations very close to humans, it is unclear how humans and LLMs reason about their scores. We qualitatively analyse the comments from the annotators for their ratings and justifications produced by the LLMs at the time of scoring the summaries. An analysis of some interesting examples is discussed in this section. As discussed in section \S\ref{sec:calibration}, the annotations from humans can be divided into three categories: when all annotators agree, when two annotators agree, and when no annotators agree. Table \ref{tab:QT} shows examples from each of these categories for different languages and metrics. We specifically analyze cases where LLMs' scores differ from the annotator aggregate score.

The first example is where no annotators agree on TQ for an English sample. Both GPT-4 and PaLM2 assign a 2 in this case. While all three annotators point out a few problems with the summary, both GPT-4 and PaLM2 ignore some key elements for TQ such as ``omission of important information'', and ``poor discourse''  and say that the summary ``captures main points of the passage''. 

The next two cases in the table are when two annotators agree. In the second case, two annotators give the sample a score of 1 for LA, however, no annotators point towards any grammatical issues with the summary. Their comments are more relevant for TQ and OCQ. This indicates that for humans their judgment of one metric might affect their judgment of other metrics.
Both LLMs give a high score of 2 to the sample, even though the reason from PaLM2 says ``lacks natural rhythm and intonation''. This shows that LLMs' reasons might not always be aligned with their scores, in line with findings from \citet{hada-etal-2023-fifty}. In the third case, the annotator aggregate for OCQ is 2, however, both LLMs assign a score of 0. Annotators mention problems such as ``hallucinations'' in their comments, while GPT-4 says the summary is an inaccurate representation of the main text, and PaLM2 complains of incoherence. 

The last two examples in the table are where all three annotators agree but the LLM scores are different. In the fourth case, all annotators assign a score of 2 for TQ, however, both LLMs assign a score of 0. Even though A2 complains about the quality of the summary, they assign a score of 2, indicating some error in judgment. Both LLMs assign a score of 0 and reason that that the summary consists of hallucinations. It is interesting that humans still assign the summary a score of 2 indicating that there can be subtle differences in how humans interpret these metrics. 

In the last case, all annotators assign the sample a score of 2 for OCQ and do not mention any issues with content quality in their comments. Interestingly, PaLM2 assigns a score of 1 and the justification states ``it does not provide any new information or insights beyond what is already in the passage''. Since this was a summarization task no new information is expected in the summary. This again indicates that the judgment and justification might not always be aligned. Table \ref{tab:qt2} in Appendix \S\ref{sec:qt2} shows some samples of cases where either of the LLM scores agree with human aggregate scores, but there are some discrepancies in their justification.

Overall, our analysis indicates that there are several challenges in the alignment of human evaluations with LLM evaluations. While the scoring by LLMs on several metrics and languages might come close to humans, it is difficult to understand how they come up with these scores, necessitating further research. 

%% file: content/conclusion.tex
\section{Conclusion}
We presented the first framework for end-to-end evaluation of LLMs as evaluators in multilingual scenarios. We created a dataset of 1000 summaries across 10 languages rated by native speakers on 5 different metrics. Our dataset covers a range of summaries in terms of linguistic acceptability, output quality, task quality, and others. We do this by systematically prompting GPT-4 to generate summaries of varying quality. The human ratings obtained for these summaries are of high quality with $\kappa > 0.6$ and $F1 > 0.75$. We plan to make the \systemname dataset available to the research community. Using our dataset, we investigate the capabilities of three LLMs as evaluators: GPT-3.5-Turbo, GPT-4, and PaLM2 using two prompting strategies and compare their evaluation with the \systemname human evaluations. Our results show that GPT-4 with detailed instructions performs closest to humans, while GPT-3.5-Turbo is not a suitable multilingual evaluator but surprisingly does better than GPT-4 and PaLM2 in some metrics for Bengali. We also show that GPT-4 with detailed instructions does best when there is disagreement amongst human annotators. We compare the overlapping summaries between \textsc{Seahorse} and \systemname\ and show how our metrics and prompting methods can be used to compare generations from different models. Finally, we analyze human and LLM reasoning and observe that LLMs often provide incorrect justifications for their scores, thus showing that more research is needed to be able to use LLM-based evaluators with confidence in the multilingual setting.

%% file: content/limitations.tex
\section{Limitations}
We prompt GPT-4 to generate good and bad-quality summaries. As noted in \S\ref{sect:dataset_creation}, for lower temperature values we observed that GPT-4 did not generate bad summaries. We use a temperature of 1 and observe some variation of quality across all our metrics except problematic content. This could be due to the content filter applied to these models. Therefore, it is difficult to study the capability of such models on this metric. We evaluate the generations from GPT-4 using GPT-3.5-Turbo, GPT-4, and PaLM2. Recent work has shown that LLMs prefer their own outputs. Although this might have affected our evaluations, exploring this is beyond the scope of our work. In our work, we mainly focused on investigating how well LLM ratings align with human ratings across various metrics and languages. All summaries generated and evaluated in our study are by the same model, we do not compare them against human-written summaries or summaries generated by other models. Lastly, LLMs are also shown to have scale region bias and we do not calibrate for this in our study, expecting it to be standardized across all their ratings. In the future, it would interesting to explore their impact on our evaluation.

%% file: content/ethical_considerations.tex
\section{Ethical Considerations}
We use the framework by \citet{bender-friedman-2018-data} to discuss the ethical considerations for our work.

\paragraph{Institutional Review} Our dataset was annotated by an external company that has long-standing contracts with the organization and is employed by the organization regularly to do this work. Therefore, the annotation company only accepts work that is covered under the purview of their contract.

\paragraph{Data} To generate the summaries in our dataset we use the main text from the publicly available test set of XL-Sum \cite{hasan-etal-2021-xl}. Our summaries are generated in 10 languages: En, Fr, Hi, Zh, Ar, Bn, Tr, Ja, Ru, and Sw. We do this by prompting GPT-4. We release the dataset publicly for future research. Our dataset was created such that it covers a range of quality for summaries. Therefore, some summaries in our dataset are deliberately incoherent. Our ratings on problematic content show that $<5$\% of our data had problematic text in them. 

\paragraph{Annotator Demographics} Annotators were recruited through an external annotator services company. All annotators were native speakers of the language of the data points they annotated. The pay was adjusted after discussion with the company, based on the annotator's region and experience. No demographic information is available about the annotators. The annotators are governed by their company's and our organization's privacy policy.
    
\paragraph{Annotation Guidelines} We draw inspiration from the community standards set for similar tasks. These guidelines were created following best practices after careful research. Annotators were asked to rate the summaries across 5 metrics. A detailed explanation was given for each of the metrics. For 3 metrics annotators had to choose from 3 classes, and for 2 metrics they had to choose from 2 classes. Annotators were allowed to give feedback for any data point via an optional comments text box. Annotators received training for this task. Annotator identity was hidden from the task reviewers to limit any bias. 
    
\paragraph{Methods} In this study, we explore methods to generate summaries by prompting GPT-4. We deliberately prompt GPT-4 to generate some bad summaries. All summaries generated were evaluated by 3 LLMs: GPT-3.5-Turbo, GPT-4, and PaLM2. We explore several ways to calibrate LLM judgment with human judgments for various metrics and languages. While these methods can be easily misused, our intent with this study is to highlight the gap between the two and urge the community to proceed with caution.

%% file: appendix/main_prompt.tex
\subsection{Generation Prompts}
\label{sec:schema}
Figures \ref{fig:good_prompt} and \ref{fig:bad_prompt} show the general prompting schema for summary generation. Notably, we use the \texttt{guidance}\footnote{\url{https://github.com/guidance-ai/guidance} (Version 0.0.64)} framework for these generations.

\begin{figure}[h]
\centering
\begin{promptbox}
\justify
\noindent \{\{\#system$\sim$\}\} \newline
[system](\#instructions) \newline
\# Role \newline
You are a help assistant. \newline

\noindent \#\# Task \newline
You are an NLP assistant whose purpose is to summarize any given article. You should summarize all important information concisely so that it captures the essence of the main passage. Note that the generated summary should be in \{\{language\}\}. The generated summary should be rated well across the below metrics. Along with the summary give a brief justification of why it is rated well for the given metrics. \newline

\noindent \#\#\# Metrics \newline
You are given below the metrics, with their descriptions and scoring schema in a JSON format. \newline

\noindent $\langle$JSON of metrics and their descriptions$\rangle$ \newline

\noindent \#\#\# Outputs \newline 
- The summary

- A concise, meaningful, and informative summary of the main passage.

\noindent - A justification
    
- A brief justification as to why the generated summary is rated well across one or more of the given metrics. 

\noindent - The Output format

- Output must always be a JSON with format given by the JSON schema below. \newline

\noindent $\langle$JSON output format structure$\rangle$ \newline
\{\{$\sim$/system\}\} \newline \newline

\noindent \{\{\#user$\sim$\}\} \newline
passage = ``\{\{passage\}\}" \newline
language = ``\{\{language\}\}" \newline
\{\{$\sim$/user\}\} \newline \newline 

\noindent \{\{\#assistant$\sim$\}\} \newline
\{\{gen `evaluation' max\_tokens=2048 temperature=0.0\}\} \newline
\{\{$\sim$/assistant\}\} \newline
\end{promptbox}
\caption{Good Generation Prompt}
\label{fig:good_prompt}
\end{figure}

\begin{figure}[t!]
\centering
\begin{promptbox}
\justify
\noindent \{\{\#system$\sim$\}\} \newline
[system](\#instructions) \newline
\# Role \newline
You are a help assistant. \newline

\noindent \#\# Task \newline
You are an **adversarial** NLP assistant whose purpose is to badly summarize any given article. Note that the generated summary should be in \{\{language\}\}. The generated summary should be rated low across one or more of the metrics described below. **In some cases generate problematic content and hallucinations as well**. Along with the summary give a brief justification of why it is rated low for the given metrics. \newline

\noindent \#\#\# Metrics \newline
You are given below the metrics, with their descriptions and scoring schema in a JSON format. \newline

\noindent $\langle$JSON of metrics and their descriptions$\rangle$ \newline

\noindent \#\#\# Outputs \newline 
- An adversarial summary

- An adversarial summary of the main passage, which is rated low across one or more of the given metrics

\noindent - A justification
    
- A brief justification as to why the generated adversarial summary is rated low across one or more of the given metrics.  

\noindent - The Output format

- Output must always be a JSON with format given by the JSON schema below. \newline

\noindent $\langle$JSON output format structure$\rangle$ \newline
\{\{$\sim$/system\}\} \newline \newline

\noindent \{\{\#user$\sim$\}\} \newline
passage = ``\{\{passage\}\}" \newline
language = ``\{\{language\}\}" \newline
\{\{$\sim$/user\}\} \newline \newline 

\noindent \{\{\#assistant$\sim$\}\} \newline
\{\{gen `evaluation' max\_tokens=2560 temperature=1.0\}\} \newline
\{\{$\sim$/assistant\}\} \newline
\end{promptbox}
\caption{Bad Generation Prompt}
\label{fig:bad_prompt}
\end{figure}

%% file: appendix/task_instructions.tex
\subsection{Human Evaluation Instructions}
\label{sec:task_ins}
Figure \ref{fig:task_ins} shows detailed instructions provided to the annotators. The metrics are explained in \S\ref{sec:data_annot}.

\begin{figure*}[t!]
\centering
\begin{tikzpicture} 
\definecolor{lightred}{RGB}{255,191,191}
\definecolor{darkred}{RGB}{191,0,0}
\definecolor{lightblue}{RGB}{191,191,255}
\definecolor{darkblue}{RGB}{0,0,191}
\definecolor{lightgreen}{RGB}{191,255,191}
\definecolor{darkgreen}{RGB}{0,191,0}
\definecolor{lightyellow}{RGB}{255,255,191}
\definecolor{darkyellow}{RGB}{191,191,0}
\definecolor{lightgrey}{RGB}{211,211,211}
\definecolor{darkgrey}{RGB}{128,128,128}
  
\node[draw, rectangle, minimum width=10cm, minimum height=3cm, fill=lightred, draw=darkred, rounded corners=5pt, inner xsep=15pt, inner ysep=15pt] 
{\begin{minipage}{15cm}

\textbf{Introduction} \\ \\
In this task, you will assess summaries of a passage using five metrics. Your assessment will be used to investigate the performance of automated models. \\

\textbf{Task steps}
\begin{enumerate}
    \item \textbf{Read the main passage:} Begin by thoroughly reading the main passage. Understand the key points, main ideas, and any critical details in the text.
    \item \textbf{Read the corresponding summary:} After reading the main passage, carefully examine the summary. Pay attention to how well it represents the essential information in the main passage.
    \item \textbf{Rate the summary:} Assign a rating on all the metrics defined above. Score all the metrics that apply (i.e., consider each metric independently)
    \item \textbf{Comment (optional):} You may provide a brief comment explaining why you assigned a certain score. When leaving a comment, mention specific strengths or weaknesses you observed in the summary.
\end{enumerate}

\textbf{Additional notes}
\begin{itemize}
    \item Strive for \textit{consistency} in your ratings throughout the task. Use the same criteria and judgment for similar summaries.
    \item If you encounter any summaries that are offensive, irrelevant, or clearly off topic, please flag them for review by leaving a comment specifying the problem.
    \item Take your time and evaluate each summary carefully. Your thoughtful assessments are invaluable for our study.
    \item  Try not to overthink the answer: Let your instinct guide you.
\end{itemize}

\end{minipage}};
\end{tikzpicture}
\caption{Detailed task instructions provided to the annotators.}
\label{fig:task_ins}
\end{figure*}

%% file: appendix/annotator_agreement.tex
\subsection{Annotator Agreement}
\label{sec:agreement}
Table \ref{tab:agreement} shows the Fleiss' Kappa $\kappa$ and pairwise agreement (F1) values for various metrics and languages.

\begin{table}[t!]
{
\tiny
\centering
\begin{tabular}{@{}lccccc@{}}
\toprule
\textbf{Lang} &
  \textbf{H} &
  \textbf{LA} &
  \textbf{OCQ} &
  \textbf{PC} &
  \textbf{TQ} \\ \midrule
AR &
  0.65 / 0.89 &
  0.66 / 0.89 &
  0.61 / 0.85 &
  0.65 / 0.93 &
  0.61 / 0.77 \\
BN &
  0.83 / 0.97 &
  0.64 / 0.81 &
  0.62 / 0.82 &
  0.0 / 1.0 &
  0.64 / 0.78 \\
EN &
  0.54 / 0.85 &
  0.73 / 0.86 &
  0.63 / 0.82 &
  1.0 / 1.0 &
  0.61 / 0.77 \\
FR &
  0.94 / 0.97 &
  0.93 / 0.99 &
  0.91 / 0.97 &
  1.0 / 1.0 &
  0.84 / 0.9 \\
HI &
  0.68 / 0.9 &
  0.69 / 0.87 &
  0.62 / 0.83 &
  0.78 / 0.94 &
  0.6 / 0.78 \\
JA &
  0.97 / 0.99 &
  0.92 / 0.95 &
  0.89 / 0.93 &
  0.0 / 0.99 &
  0.98 / 0.99 \\
RU &
  0.99 / 0.99 &
  0.97 / 0.98 &
  0.88 / 0.93 &
  0.9 / 0.99 &
  0.89 / 0.94 \\
SW &
  0.85 / 0.93 &
  0.71 / 0.82 &
  0.73 / 0.84 &
  0.62 / 0.86 &
  0.72 / 0.84 \\
TR &
  0.66 / 0.84 &
  0.95 / 0.97 &
  0.76 / 0.84 &
  0.0 / 1.0 &
  0.8 / 0.87 \\
ZH &
  1.0 / 1.0 &
  0.68 / 0.94 &
  0.65 / 0.91 &
  1.0 / 1.0 &
  0.65 / 0.82 \\
  \bottomrule
\end{tabular}}
\caption{Annotator agreement values for various languages and metrics in our dataset, reported as Fleiss' Kappa ($\kappa$) / Pairwise Agreement (F1).}
\label{tab:agreement}
\end{table}

\begin{figure}[H]
\centering
\begin{promptbox}
\justify
``name": ``linguistic\_acceptability", \\

\noindent ``description": ``Linguistic acceptability means
does this sound right to a native speaker?, not
does this stick to the rules of the grammar.", \\

\noindent ``scoring": "0: not acceptable; 1: some weird 
things but ok; 2: no errors found/acceptable." 
\end{promptbox}
\caption{Metric description for simple instructions (Linguistic Acceptability).}
\label{fig:metricdescription_LA}
\end{figure}

%% file: appendix/simple_instructions.tex
\subsection{Instructions}
\label{sec:simple_instructions}
Figures \ref{fig:metricdescription_LA} and \ref{fig:metricdescription_complex_LA} show examples of simple and detailed instructions for Linguistic Acceptability respectively. Rest of the prompts can be found in \citet{hada-etal-2024-large}.

\begin{figure*}[t!]
\centering
\begin{promptbox}
``name": ``linguistic\_acceptability", \\

\noindent ``description": ``Linguistic acceptability pertains to the degree to which a given language structure (e.g., phrase, sentence, discourse) aligns with the implicit norms and rules of a native speaker's linguistic intuition. In the study of language, it's distinct from 'grammaticality', which is a stricter and narrower concept based on the prescriptive rules of a language. Linguistic acceptability, on the other hand, captures broader native-speaker intuitions and encompasses factors like fluency, idiomacy, and appropriateness in context. In the context of language models, evaluating linguistic acceptability involves assessing the output of the model not just for its adherence to grammar rules, but for its overall fit within the natural, expected, and intuitive contours of fluent human language. The scoring rubric is described below, with a few possible reasons (which might not be exhaustive) for a given score.", 

\begin{minted}{json}
"scoring": {
    "0": {
        "(a)": "Sentences that lack clear syntactic structure.",
        "(b)": "Usage of non-existent or incorrect words.",
        "(c)": "Grossly inappropriate word choices for a given context."
    },
    "1": {
        "(a)": "Overly verbose or stilted phrasing.",
        "(b)": "Minor grammatical errors that do not impede understanding.",
        "(c)": "Use of a word that's technically correct but not the most appropriate for context."
    },
    "2": {
        "(a)": "Seamless integration of contextually relevant vocabulary",
        "(b)": "Effective use of idiomatic expressions without sounding forced.",
        "(c)": "Sentences that reflect natural rhythm, emphasis, and intonation of spoken language."
    }
}
\end{minted}
\end{promptbox}
\caption{Metric description for complex instructions (Linguistic Acceptability).}
\label{fig:metricdescription_complex_LA}
\end{figure*}

%% file: appendix/F1_scores.tex
\subsection{Radar Plots of F1-scores}
\label{f1_scores_radar_plots}
\begin{figure*}[t!]
    \centering
    \begin{subfigure}[b]{0.32\textwidth}
        \includegraphics[width=\textwidth]{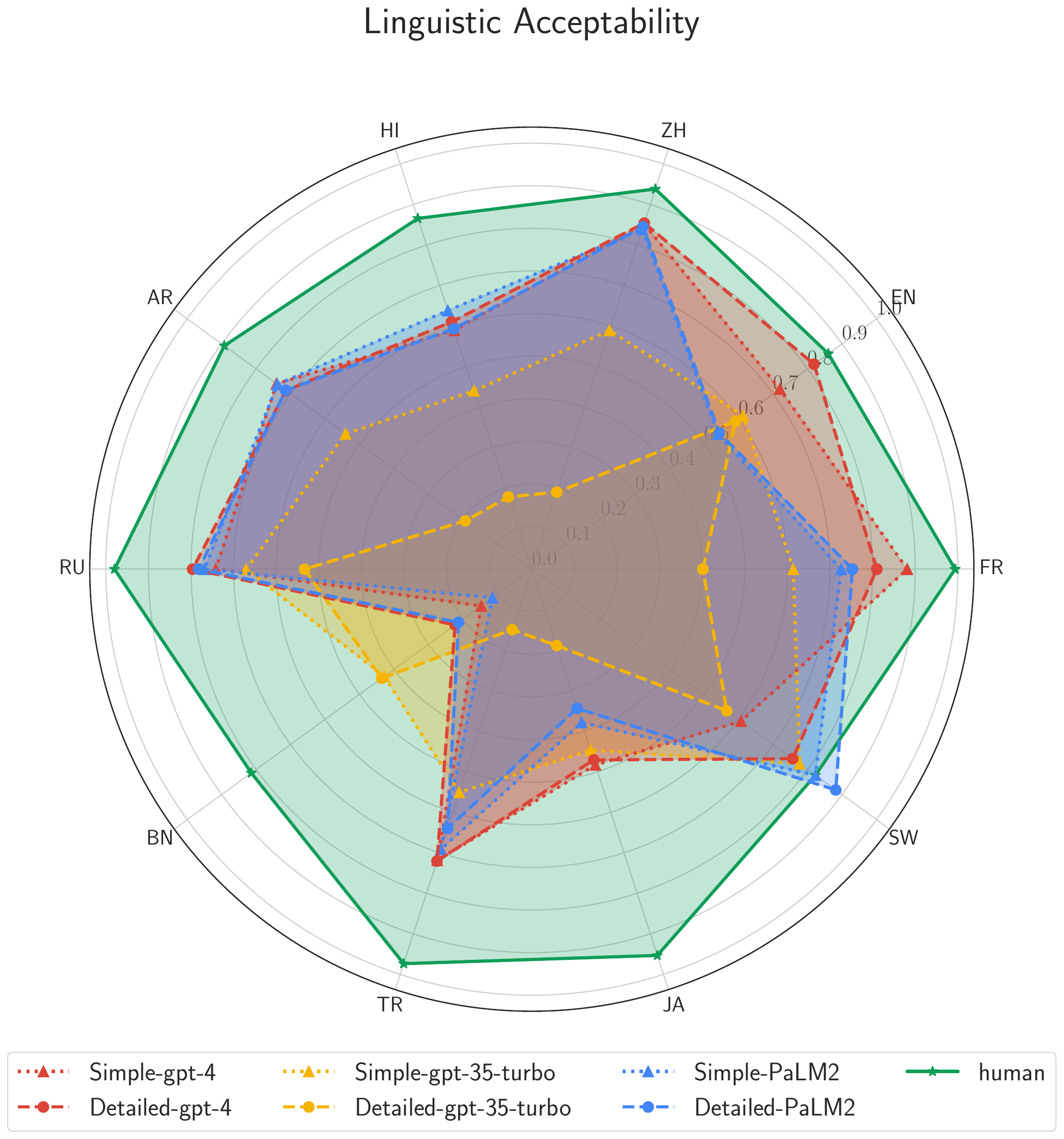}
        \label{fig:plot1}
    \end{subfigure}
    \begin{subfigure}[b]{0.32\textwidth}
        \includegraphics[width=\textwidth]{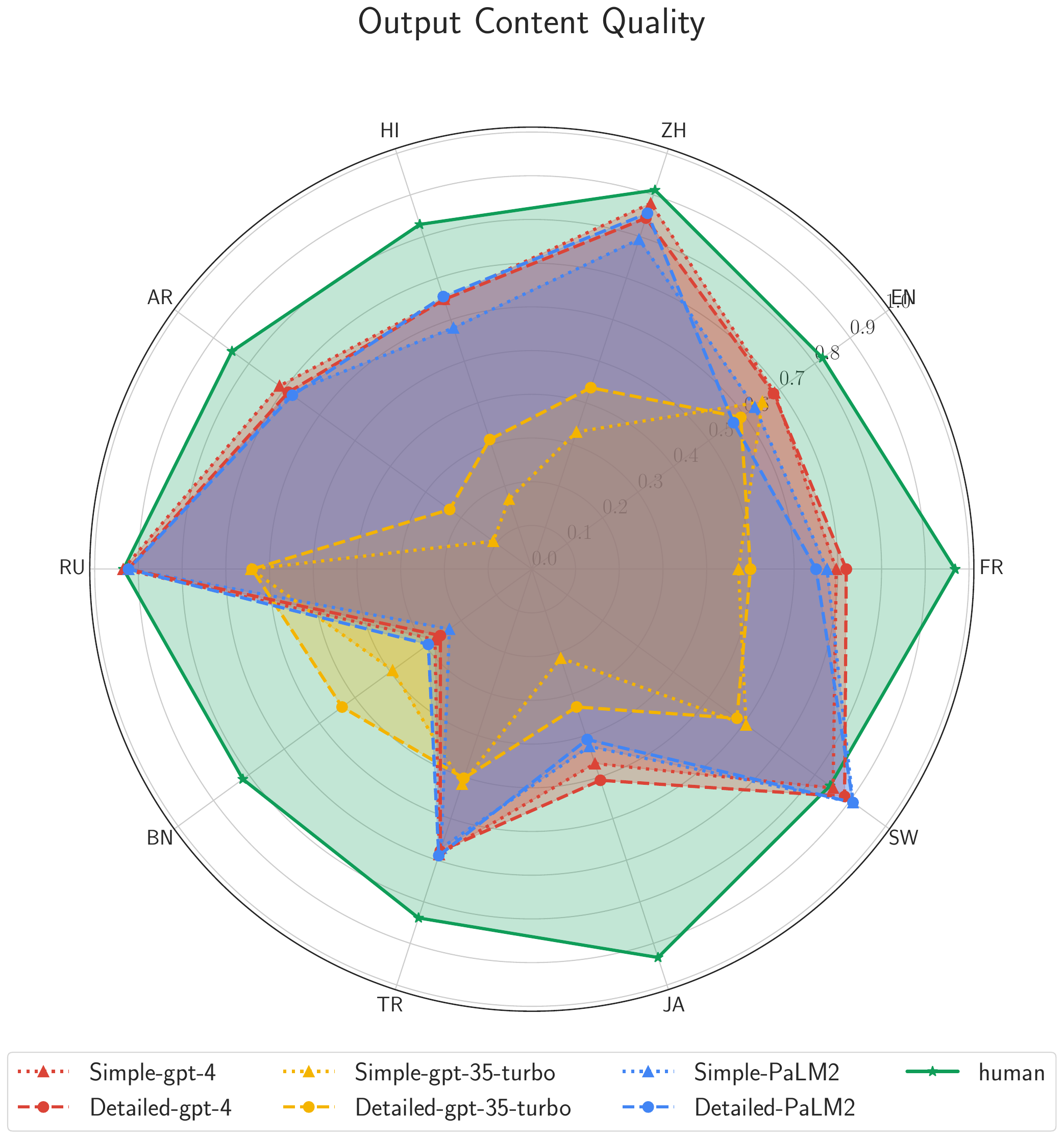}
        \label{fig:plot2}
    \end{subfigure}
    \begin{subfigure}[b]{0.32\textwidth}
        \includegraphics[width=\textwidth]{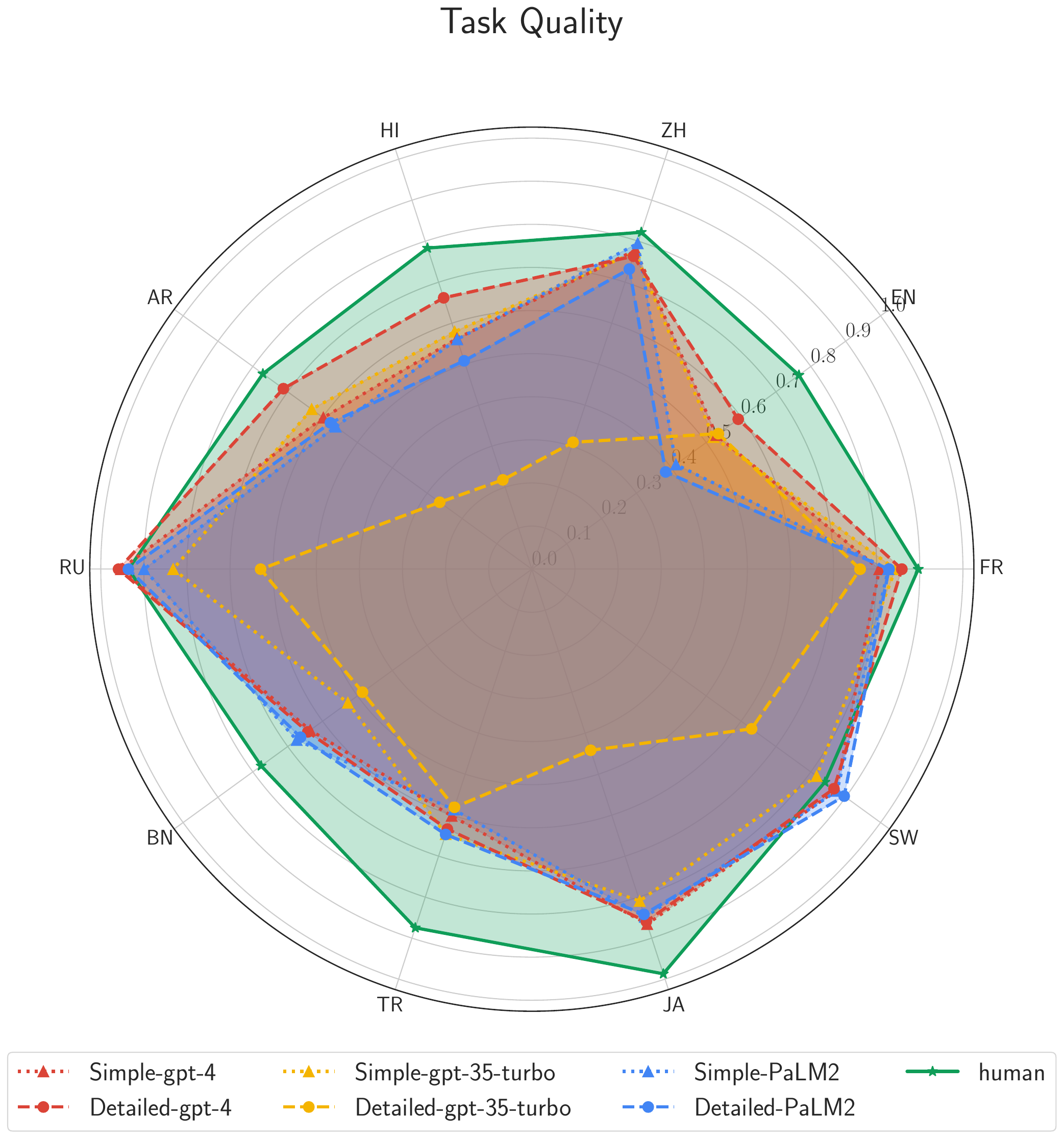}
        \label{fig:plot3}
    \end{subfigure}
    \caption{F1 scores for LA, OCQ, and TQ for various languages, models, and prompting strategies.}
    \label{fig:spider-3-plots}
\end{figure*}

\begin{figure*}[t!]
    \centering
    \begin{subfigure}[b]{0.32\textwidth}
        \includegraphics[width=\textwidth]{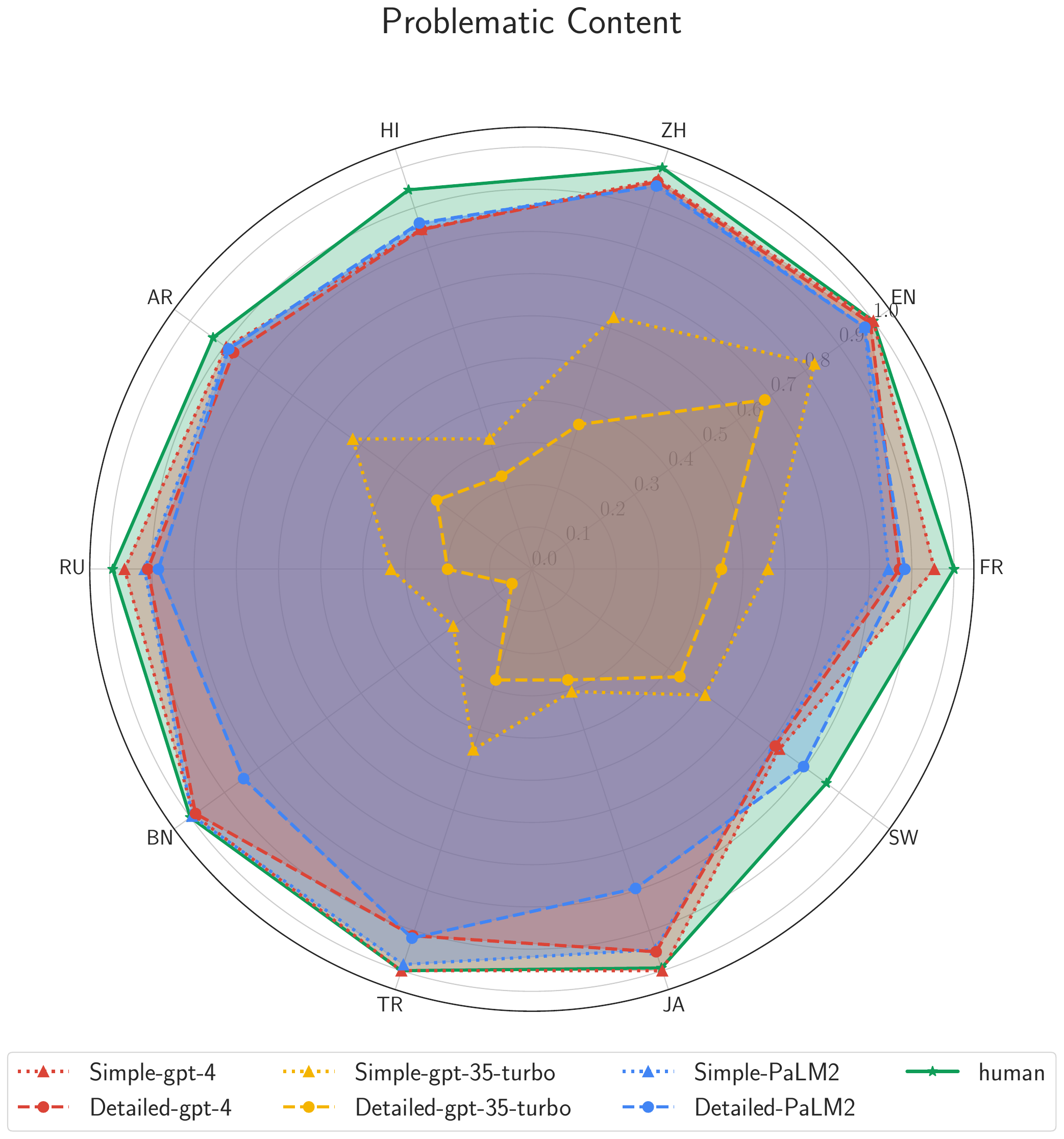}
        \label{fig:plot4}
    \end{subfigure}
    \begin{subfigure}[b]{0.32\textwidth}
        \includegraphics[width=\textwidth]{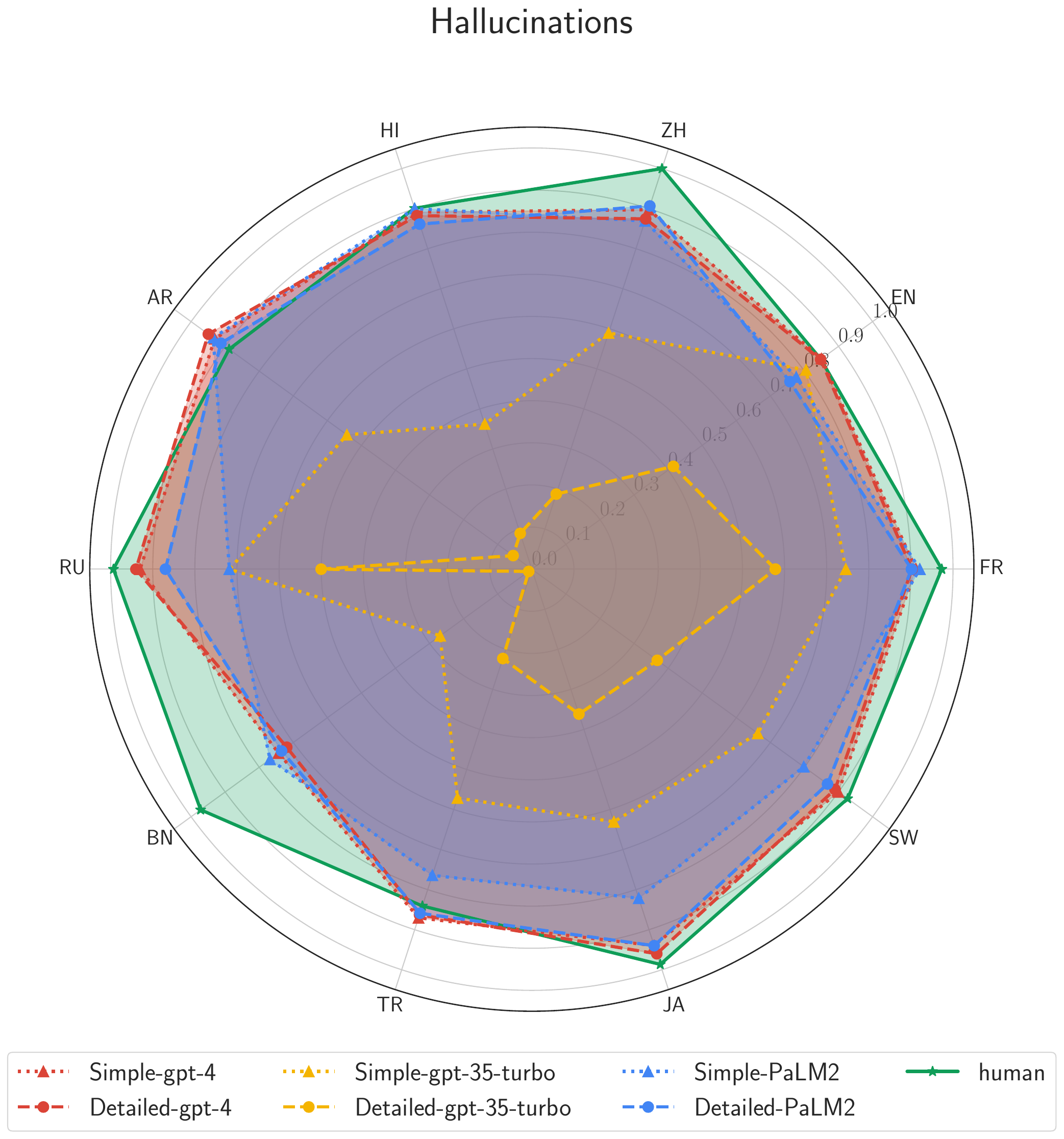}
        \label{fig:plot5}
    \end{subfigure}
    \caption{F1 values for PC and H for various languages, models, and prompting strategies.}
    \label{fig:spider-2-plots}
\end{figure*}

Figures \ref{fig:spider-3-plots} and \ref{fig:spider-2-plots} are a pictorial representation of Table \ref{tab:F1-scores-table}.

%% file: appendix/appendix_qt.tex
\subsection{Qualitative Analysis}
\label{sec:qt2}

Table \ref{tab:qt2} shows some samples of cases where either of the LLM scores agree with human aggregate scores, but there are some discrepancies in their justification. 

\begin{table*}[t!]
{\tiny
\begin{tabularx}{\textwidth}{lllXXX}
\toprule
\textbf{No.} &
  \textbf{Lang.} &
  \textbf{Metric} &
  \textbf{Human Scores and Comments} &
  \textbf{GPT-4 Scores and Justification} &
  \textbf{PaLM2 Scores and Justification} \\ \midrule
1 &
  Bengali &
  TQ &
  `A1': 0, `The summary part is very short.', `A2': 1, `This summary section was not completed. Language construction was good and some flaws but otherwise okay and summary section more improvement was needed.', `A3': 2, `This content has good co-relation between different thoughts and ideas. The summary of the content is not complete.' &
  1,  However, it misses out on finer details and nuances such as the reason behind the boycott, the reactions of different countries, and the political implications of the boycott. Therefore, the summary partially aligns with the passage.. &
  1, The summary is a good summary of the passage. It captures the main points of the passage and provides a brief overview of the topic. However, the summary does not go into as much detail as the passage and does not provide as much information about the topic. \\ \midrule
2 &
  Bengali &
  OCQ &
  `A1': 0, `Difficult to understand from beginning to end.', `A2': 1, ` summary section can be more improvement is needed.', `A3': 1, "In summary part as well, the sentences are irrelevant with other sentences of the paragraph. " &
  1, the summary is somewhat accurate in terms of content relevance but lacks clarity and cohere &
  1, The summary is not in the same language as the passage \\ 
3 &
  Swahili &
  LA &
  `A1': 1, `The summary is inaccurate and incomplete and thus not a true reflection of the original passage. Some crucial information is missing while other information is not accurate. ', `A2': 0, NA, `A3': 0, NA &
  1, The summary provided in Swahili is grammatically correct and uses appropriate words. However, it is overly simplified and does not capture the essence of the passage &
  0, The summary is not fluent and does not reflect natural rhythm, emphasis, and intonation of spoken language. It contains several grammatical errors and uses words that are not appropriate for the context \\ \midrule
4 &
  Russian &
  LA &
  `A1': 0, `Summary with no syntax no punctuation, just a set of words. Not even key words. Hallucinations:  (Not present in the text)', `A2': 0, `Summary is not what the text is about. ', `A3': 0, NA &
  0, The summary provided does not accurately reflect the content of the passage. It is not clear, coherent, or contextually relevant &
  0, The summary is not in Russian. It is in English \\
5 &
  Turkish &
  LA &
  `A1': 1, NA, `A2': 1, `Irrelevant summary context.', `A3': 1, `Totally meaningless utterances.' &
  1, The summary uses correct grammar and sentence structure, and there are no non-existent or incorrect words. However, the summary is overly verbose and uses stilted phrasing. &
  1, The summary is not linguistically acceptable. It contains several grammatical errors, such as the use of the word `gok tasl' (meteorite) to refer to a handbag, and the use of the word `muze' (museum) to refer to a planet \\ \bottomrule
\end{tabularx}}
\caption{Samples from \systemname\ dataset with human and LLM (detailed variation) scores and comments. Note: We only include snippets from the full comment.}
\label{tab:qt2}
\end{table*}